\documentclass[letterpaper,twocolumn]{article}

\usepackage{algpseudocode}
\usepackage{algorithm}
\usepackage{amsfonts}
\usepackage{amsmath}
\usepackage{amssymb}
\usepackage{amstext}
\usepackage{amsthm}
\usepackage[toc,page]{appendix}
\usepackage{array}
\usepackage{authblk}
\usepackage{booktabs}
\usepackage{cite}
\usepackage{datetime}
\usepackage{enumitem}
\usepackage{etoolbox}
\usepackage[T1]{fontenc}
\usepackage{framed}
\usepackage{fullpage}
\usepackage[margin=0.7in]{geometry}
\usepackage{graphics}
\usepackage{graphicx}
\usepackage[utf8]{inputenc}
\usepackage{mathrsfs}
\usepackage{microtype}
\usepackage{multirow}
\usepackage{nicefrac}
\usepackage{pdflscape}
\usepackage{pgfplots}
\usepackage{rotating}
\usepackage{setspace}
\usepackage[position=b]{subcaption}
\usepackage{tikz}
\usepackage{times}
\usepackage{units}
\usepackage{url}
\usepackage{xcolor}
\usepackage[font=small]{caption}
\usepackage[framemethod=TikZ]{mdframed}
\usepackage{hyperref}

\usetikzlibrary{shapes,arrows,plotmarks,matrix,positioning,fit,calc,3d,patterns,decorations.pathreplacing,intersections}

\newmdenv[
roundcorner=10pt,
backgroundcolor=gray!10,
linecolor=gray!10,
tikzsetting={draw=black,line width=3pt,dashed,dash pattern= on 10pt off 3pt},
outerlinewidth=0pt,
innerlinewidth=0pt,
]{figbox}

\newlength\figureheight
\newlength\figurewidth

\newcommand{\iid}{\stackrel{\mathrm{iid}}{\sim}}
\newtheorem{theorem}{Theorem}[section]

\newtheorem{proposition}[theorem]{Proposition}

\newcounter{num}
\title{A Transportation $L^p$ Distance for Signal Analysis}

\author[1]{Matthew Thorpe\footnote{mthorpe@andrew.cmu.edu}}
\author[1]{Serim Park\footnote{serimp@andrew.cmu.edu}}
\author[3]{Soheil Kolouri\footnote{skolouri@hrl.com}}
\author[2]{Gustavo K. Rohde\footnote{gustavo@virginia.edu}}
\author[1]{Dejan Slep\v{c}ev\footnote{slepcev@math.cmu.edu}}
\affil[1]{Carnegie Mellon University, Pittsburgh, PA 15213, USA}
\affil[2]{University of Virginia, Charlottesville, VA 22908, USA}
\affil[3]{HRL Laboratories, Malibu, CA 90265, USA}
\date{September 2016}

\begin{document}

\maketitle

\begin{abstract}
Transport based distances, such as the Wasserstein distance and earth mover's distance, have been shown to be an effective tool in signal and image analysis. 
The success of transport based distances is in part due to their Lagrangian nature which allows it to capture the important variations in many signal classes. 
However these distances require the signal to be nonnegative and normalized. Furthermore, the signals are considered as measures and compared by redistributing (transporting) them, which does not directly take into account the signal intensity.
Here we  study a transport-based distance, called the $TL^p$ distance, that combines Lagrangian and intensity modelling and is directly applicable to general, non-positive and multi-channelled signals.
The framework allows the application of existing numerical methods.
We give an overview of the basic properties of this distance and applications to classification, with multi-channelled, non-positive one and two-dimensional signals, and color transfer.
\end{abstract}

\section{Introduction \label{sec:Intro}}

Enabled by advances in numerical implementation \cite{benamou15,cuturi13,oberman15}, and their Lagrangian nature, transportation based distances for signal analysis are becoming increasingly popular in a large range of applications.
Recent applications include astronomy \cite{brenier03,frisch02,frisch04}, biomedical sciences \cite{basu14,haker03,haker01,haker04,su15,tosun15,urrehman09,zhu05,zhu07}, colour transfer \cite{chizat16,ferradans13,morovic03,rabin14,rabin15}, computer vision and graphics \cite{bonneel15,lipman11,pele09,rabin10,rubner00,solomon15,solomon14a}, imaging \cite{kolouri15,lellmann14,rabin11}, information theory \cite{tannenbaum10}, machine learning \cite{astrom16,courty14,frogner15,kolouri16b,kolouri16a,montavon15,solomon14b}, operational research \cite{russell69} and signal processing \cite{oudre12,park15}.

The success of transport based distances is due to the large number of applications that consider signals that are Lagrangian in nature (spatial rearrangements, i.e. transport, are a key factor when considering image differences).
Many signals contain similar features for which transport based distances will outperform distances that only consider differences in intensity, such as $L^p$.
Optimal transport (OT) distances, for example the earth mover's distance or Wasserstein distance, are examples of transport distances.
However these distances do not directly account for signal intensity.
The $L^p$ distance is the other extreme, this distance is based on intensity and does not take into account Lagrangian properties.

In this paper we develop the $TL^p$ distance introduced in~\cite{garciatrillos15} which combines both Lagrangian and intensity based modelling.
Our aim is to show that by including both transport and intensity within the distance we can better represent the similarities between classes of data in many problems.  
For example, if a distance can naturally differentiate between classes, that is the within class distance is small compared to the between class separation, then the classification problem is made easier.
This requires designing distances that can faithfully represent the structure within a given data set.

Optimal transport distances interpret signals as either probability measures or as densities of probability measures.
This places restrictions on the type of signals one can consider.
Probability measures must be non-negative, integrate to unity and be single-channelled.
In order to apply OT to a wider class of signals one has to use ad-hoc methods to transform the signal into a probability measure.
This can often dampen the features, for example renormalization may reduce the intensity range of a signal.

The $TL^p$ distance does not need the signal to be a probability measure and therefore the above restrictions do not apply.
Rather, the $TL^p$ distance models the intensity directly.
The framework is sufficiently general as to include signals on either a discrete or continuous domains that can be negative, multi-channelled and integrate to an arbitrary value.

Another property of OT, due to the lack of intensity modelling, is its insensitivity to high frequency perturbations.
This is due to transport being on the order of the wavelength of the perturbation.
By modelling the intensity directly, and therefore accounting for amplitude, the $TL^p$ distance does not suffer this property.

The aim of this paper is to develop the $TL^p$ framework and demonstrate its applicability in a range of applications.
We consider classification problems on data sets where we show that the $TL^p$ better represents the underlying geometry, i.e. achieves a better between class to within class distance, than popular alternative distances.

We also consider the colour transfer problem in a context where spatial information, as well as intensity, is important.
To apply standardised tests in applications such as medical imaging it is often necessary to normalise colour variation~\cite{khan14,magee09,shinohara14}.
One solution is to match the means and variance of each colour channel (in some colourspace e.g. RGB or LAB).
However, by transferring the colour of one image onto the other it is possible to recolour an image with \emph{exactly} the same colour profile.

A popular method is to use the OT distance on the histogram of images~\cite{chizat16,ferradans13,morovic03,rabin14,rabin15}.
This allows one to take into account the intensity of pixels but includes no spatial information.
The $TL^p$ distance is able to include both spatial and intensity information.

Our methodology, therefore, has more in common with registration methods that aim to find a transformation that maximizes the similarity between two images where our measure of similarity includes both spatial and intensity information.
One should compare our approach to~\cite{haker04} where the authors develop a numerical method for the Monge formulation of OT with the addition of an intensity term for image warping and registration.
However, unlike the method presented in~\cite{haker04}, the formulation presented here defines a metric.

\paragraph*{Paper Overview.}
The outline for this paper is the following.
In the next section we review OT and give a formal definition of the $TL^p$ distance followed by examples to illustrate its features and to compare with the OT and $L^p$ distances.
In Section~\ref{sec:Frame} we give a more general definition and explain some of its key properties.
In Section~\ref{sec:App} we include applications of the $TL^p$ distances.
We first consider classification on synthetic one and two dimensional, non-positive signals with no assumption on total mass and to real-world multivariate signals and two-dimensional images.
A further application to the colour transfer problem is then given.
Conclusions are given in Section~\ref{sec:conc}.

\section{Formal Definitions and Examples \label{sec:OTandTLp}}

\subsection{Review of Optimal Transport and the \texorpdfstring{$TL^p$}{TLp} Distance \label{subsec:OTandTLP:Rev}}

We begin by reviewing optimal transport in first the Kantorovich formulation and then the Monge formulation.

\begin{figure*}[ht!]
\begin{figbox}
\centering
\setlength\figureheight{4cm}
\setlength\figurewidth{0.47\columnwidth}
\scriptsize
\begin{subfigure}{0.48\columnwidth}
\centering
\scriptsize
%
%
%
%
\begin{tikzpicture}

\begin{axis}[%
width=\figurewidth,
height=\figureheight,
axis on top,
scale only axis,
axis equal,
xmin=0,
xmax=1536,
ymin=0,
ymax=512,
hide axis,
name=plot1
]
\addplot graphics [xmin=0,xmax=512,ymin=0,ymax=512] {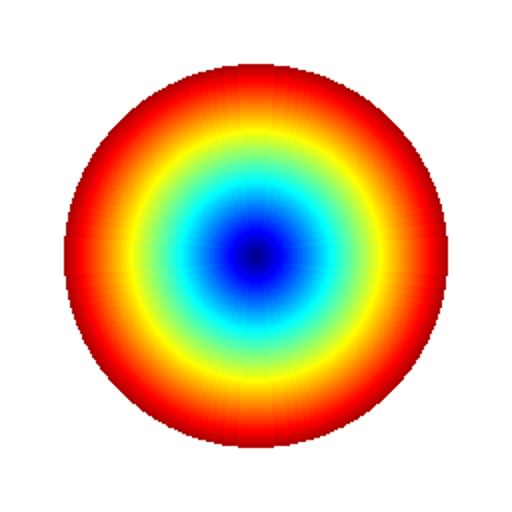};
\addplot graphics [xmin=1024,xmax=1536,ymin=0,ymax=512] {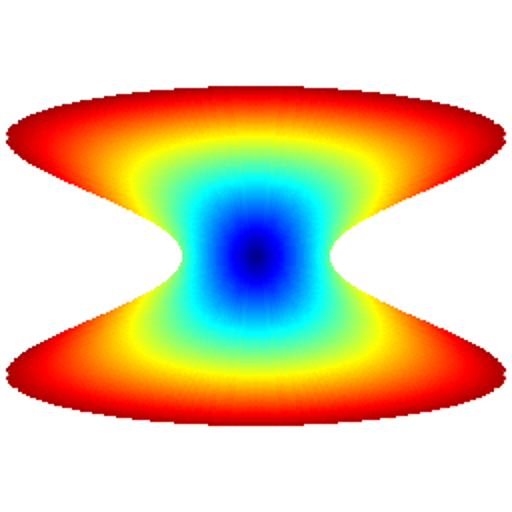};
\draw (axis cs:512,400) edge[->,bend left,thick] node [midway,above] {$T$} (axis cs:1024,400);
\draw (axis cs:64,32) node {$\mu$};
\draw (axis cs:1472,32) node {$\nu$};
\end{axis}
\end{tikzpicture}%
\caption{
The Monge formulation of OT looks for maps that rearrange $\mu$ into $\nu$.
The optimal transport map between the uniform measure on the circle and the uniform measure on a bean (with the same mass) is given above.
The mapping is represented by the colouring. 
}
\label{subfig:OTandTLp:Rev:OptMap}
\end{subfigure}
\hfill
\begin{subfigure}{0.48\columnwidth}
\centering
\scriptsize
\input{PushforwardMeasure.tikz}
\caption{
The measure $T_{\#}\mu$ is represented with support on the $y$-axis and the measure $\mu$ on the $x$-axis in the above graph.
The pushforward measure $T_{\#}\mu(A)$ is, by definition, equal to $\mu(T^{-1}(A))$.
When $T$ is differentiable, one-to-one and $\mathrm{d} \mu(x) = \rho_\mu(x) \; \mathrm{d} x$, $\mathrm{d} \nu(y) = \rho_\nu(y) \; \mathrm{d} y$ then $\nu = T_{\#}\mu$ is equivalent to $\mathrm{det} (DT(x)) \rho_\nu(T(x)) = \rho_\mu(x)$.
}
\label{subfig:OTandTLp:Rev:Pushforward}
\end{subfigure}
\caption{
On the left an example optimal transport map for OT, on the right an illustration of the pushforward measure.
}
\label{fig:OTandTLp:Rev:PushAndOpt}
\end{figbox}
\end{figure*}

\paragraph*{The Kantorovich Formulation of Optimal Transport.}
For measures $\mu$ and $\nu$ on $\Omega\subset \mathbb{R}^d$ with the same mass and a continuous cost function $c:\Omega\times \Omega\to [0,\infty)$ the Kantorovich formulation of OT is given by
\begin{equation} \label{eq:OTandTLp:Rev:KOT}
\mathrm{OT}(\mu,\nu) = \min_\pi \int_{\Omega\times \Omega} c(x,y) \; \mathrm{d} \pi(x,y)
\end{equation}
where the minimum is taken over probability measures $\pi$ on $\Omega\times\Omega$ such that the first marginal is $\mu$ and the second marginal is $\nu$, i.e. $\pi(A\times \Omega) = \mu(A)$ and $\pi(\Omega\times B) = \nu(B)$ for all open sets $A$ and $B$. 
We denote the set of such $\pi$ by $\Pi(\mu,\nu)$.
We call measures $\pi\in \Pi(\mu,\nu)$ transport plans since $\pi(A\times B)$ is the amount of mass in $A$ that is transferred to $B$. 
Minimizers $\pi^*$ of $\mathrm{OT}(\mu,\nu)$, which we call optimal plans, exist when $c$ is lower semi-continuous~\cite{villani03}.

A common choice is $c(x,y)=|x-y|^p_p=\sum_{i=1}^d|x_i-y_i|^p$ in which case we define $d_{\mathrm{OT}}(\mu,\nu)=\sqrt[p]{\mathrm{OT}(\mu,\nu)}$.
When $p=2$ this is known as the Wasserstein distance and when $p=1$ the earth mover's distance.
With an abuse of notation we will sometimes write $d_{\mathrm{OT}}(f,g)$ when $\mu$ and $\nu$ have densities $f$ and $g$ respectively.

When $\mu$ has a continuous density then the support of any optimal plan $\pi^*$ is contained on the graph of a function $T^*$.
In particular this implies $\pi^*(A,B) = \mu\left(\left\{x : x\in A, T^*(x)\in B \right\} \right)$ and furthermore that the optimal plan defines a mapping between $\mu$ and $\nu$, see for example Figure~\ref{subfig:OTandTLp:Rev:OptMap}. 
This leads us to the Monge formulation of OT.

\paragraph*{The Monge Formulation of Optimal Transport.}
An appealing property of optimal transport distances are their formulation in a Lagrangian setting.
One can rewrite the optimal transport problem in the Monge formulation as
\begin{equation} \label{eq:OTandTLp:Rev:MOT}
\mathrm{OT}_M(\mu,\nu) = \inf_T \int_{\Omega\times \Omega} c(x,T(x)) \; \mathrm{d} \mu(x)
\end{equation}
where the infimum is taken over transport maps $T:\Omega\to\Omega$ that rearrange $\mu$ into $\nu$, i.e. $\nu = T_{\#} \mu$ where we define the pushforward of $\mu$ onto the range of $T$ by $T_{\#}\mu(A) = \mu(T^{-1}(A))$, see Figure~\ref{subfig:OTandTLp:Rev:Pushforward}.
This is now a non-convex optimization problem with nonlinear constraints.
However when, for example, $\mu$ and $\nu$ have densities, then optimal transport maps $T^*$ exist and give a natural interpolation between two measures.
In particular when $c(x,y)=|x-y|^p$ the map $T_t(x) = (1-t)x + tT^*(x)$ describes the path of particle $x$ and furthermore the measure of $\mu$ pushed forward by $T_t$ is the geodesic (shortest path) between $\mu$ and $\nu$.
This property has had many uses in transport based morphometry applications such as biomedical~\cite{basu14,ozolek14,tosun14,wang12}, super-resolution~\cite{kolouri15} and has much in common with large deformation diffeomorphism techniques in shape analysis~\cite{grenander98,joshi00}.

\paragraph*{Optimal Transport in Signal and Image Processing.}
To further motivate our development of the $TL^p$ distance we point out some features of optimal transport important to signal and image processing.
We refer to~\cite{kolouri16} and references therein for more details and a review of the subject.

Key to the success of OT is the ability to provide generative models which accurately represent various families of data distributions.
The success and appeal of OT owes to (1) ability to capture well the signal variations due to spatial rearrangements (shifts, translations, transport), (2) the OT distances are theoretically well understood and have appealing features (for example Wasserstein distance has a Riemannian structure and geodesics can be characterized), (3) efficiency and accuracy of numerical methods, (4) simplicity compared to other Lagrangian methods such as large deformation diffeomorphic metric mapping. 

The Monge formulation of OT defines a mapping between images which has been used in, for example, \emph{image registration}~\cite{haber10,haker03,haker01,haker04,museyko09,urrehman09,zhu07} where one wishes to find a common geometric reference frame between two or more images.
In addition to the properties listed above the success of OT is due to the fact that (5) the Monge problem is symmetric (i.e. if $T$ is the optimal map from the first image to the second, then $T^{-1}$ is the optimal map from the second image to the first) and (6) OT provides a landmark-free and parameter-free registration scheme.

We now introduce the $TL^p$ distance in the simplest setting.

\paragraph*{The Transportation $L^p$ Distance.}
In this paper we use the $TL^p$ distance (introduced in more generality in the next section), for functions $f,g:\Omega\to\mathbb{R}^m$ defined by
\[ d_{TL^p}^p(f,g) = \min_\pi \int_{\Omega\times\Omega} |x-y|_p^p + |f(x)-g(y)|_p^p \; \mathrm{d} \pi(x,y) \]
where the minimum is taken over all probability measures $\pi$ on $\Omega\times \Omega$ such that both the marginals are the Lebesgue measure $\mathcal{L}$ on $\Omega$, i.e. $\pi \in \Pi(\mathcal{L},\mathcal{L})$.
This can be understood in two ways.
The first is as an optimal transport distance of the Lebesgue measure with itself and cost $c(x,y)=|x-y|^p_p+|f(x)-g(y)|_p^p$.
This observation allows one to apply existing numerical methods for OT where the effective dimension is $d$ (recall that $\Omega \subseteq \mathbb{R}^d$).
The second is as an OT distance between the Lebesgue measure raised onto the graphs of $f$ and $g$.
That is, given $f,g:\Omega \to \mathbb{R}$ then we define the measures $\tilde{\mu},\tilde{\nu}$ on the graphs of $f$ and $g$ by $\tilde{\mu}(A\times B) = \mathcal{L}\left(\left\{ x \, : \, x \in A, f(x) \in B\right\} \right)$ and $\tilde{\nu}(A\times B) = \mathcal{L}\left(\left\{ y \, : \, y \in A, g(y) \in B\right\} \right)$ for any open sets $A\subseteq \Omega, B \subseteq \mathbb{R}^m$.
The $TL^p$ distance between $f$ and $g$ is the OT distance between $\tilde{\mu}$ and $\tilde{\nu}$.
Transport in $TL^p$ is of the form $(x,f(x))\mapsto (y,g(y))$ and therefore has two components.
We refer to horizontal transport as the transport $x\mapsto y$ in $\Omega$, and vertical transport as the transport $f(x)\mapsto g(y)$.
In the next section we discuss the behaviour of $TL^p$ through three examples.

\subsection{Examples Illustrating the Behaviour of \texorpdfstring{$TL^p$}{TLp} \label{subsec:OTandTLp:Ex}}

\paragraph*{No mass renormalization.}
Unlike for OT, in the $TL^p$ distance there is no need to assume that $f$ and $g$ are non-negative or that they have the same mass.
If a signal is negative then a typical (ad-hoc) fix in OT is to add a constant to make the signal non-negative before computing the distance.
How to choose this constant is often unclear unless a lower bound is known a-priori.
Furthermore this may damage sensitivity to translations as the defining features of the signal become compressed.
For example, considering the functions in Figure~\ref{subfig:OTandTLp:Rev:TL2vOT:TL2ReNorm}, let $g = f(\cdot-\ell)$ be the translation of $f$.
OT will lose sensitivity when comparing $\hat{f}=\frac{f+\alpha}{\int (f + \alpha)}$ and $\hat{g}=\frac{g+\alpha}{\int (g + \alpha)}$.
In particular $d_{\mathrm{OT}}(\hat{f},\hat{g})$ scales with the height of the renormalised function, which is of the order of $\frac{1}{\alpha}$, and the size of the shift: $d_{\mathrm{OT}}(\hat{f},\hat{g}) \propto \frac{h_0\ell}{\alpha}$ where $h_0$ is the height of $f$.
To ensure positivity one must choose $\alpha$ large but this also implies a small OT distance.
Note also that both $L^p$ and $TL^p$ are invariant under adding a constant whereas OT is not.

\paragraph*{Sensitivity to High Frequency Perturbations.}
The $TL^p$ distance inherits sensitivity to high frequency perturbations from the $L^p$ norm.
For example, let $g=f+A\xi$ where $\xi$ is high frequency perturbation with amplitude $A$ and wavelength $\omega$.
Then the distance moved by each particle in the Monge formulation of OT is on the order of the wavelength $\omega$ of $\xi$, which is small, and independent of the amplitude $A$.
On the other hand both the $TL^p$ distance and the $L^p$ distance are independent of the wavelength but scale linearly with amplitude, see Figure~\ref{subfig:OTandTLp:Rev:TL2vOT:TL2HighFreq}.
In particular OT is insensitive to high frequency noise regardless of how large the amplitude whereas both $TL^p$ and $L^p$ scale linearly with the amplitude.

\begin{figure*}[ht]
\begin{figbox}
\centering
\setlength\figureheight{3.5cm}
\setlength\figurewidth{0.45\columnwidth}
\scriptsize
\begin{subfigure}{0.45\columnwidth}
\centering
\scriptsize
\begin{tikzpicture}

\begin{axis}[%
width=\figurewidth,
height=\figureheight,
scale only axis,
xmin=-0.2,
xmax=3.2,
ymin=-6,
ymax=10,
hide axis,
name=plot1,
axis background/.style={fill=white!100},
at={(0,0)}
]
\draw [->] (axis cs: 0,0) -- (axis cs: 3,0);
\draw [->] (axis cs: 0,-5) -- (axis cs: 0,9);
\draw node at (axis cs:-0.1,4.1) {1};
\draw [line width=1.2pt] (axis cs: 0,0) -- (axis cs: 0.1,0);
\addplot+[no marks,black,line width=1.2pt,domain=0.1:1.1,samples=100] {0.75*(sin(540*((x)-0.6)) + sin(180*((x)-0.6)))};
\addplot+[no marks,black,line width=1.2pt,domain=1.1:2.1,samples=100] {1.25*(sin(540*((x)-0.6)) + sin(180*((x)-0.6)))};
\draw [line width=1.2pt] (axis cs: 2.1,0) -- (axis cs: 2.8,0);
\draw [line width=1.2pt,red] (axis cs: 0,0.05) -- (axis cs: 0.3,0.05);
\addplot+[no marks,red,line width=1.2pt,domain=0.3:1.3,samples=100] {0.75*(sin(540*((x)-0.8)) + sin(180*((x)-0.8))) + 0.05};
\addplot+[no marks,red,line width=1.2pt,domain=1.3:2.3,samples=100] {1.25*(sin(540*((x)-0.8)) + sin(180*((x)-0.8))) + 0.05};
\draw [line width=1.2pt,red] (axis cs: 2.3,0.05) -- (axis cs: 2.8,0.05);
\draw [line width=1.2pt,dashed] (axis cs: 0,4.1) -- (axis cs: 0.1,4.1);
\addplot+[no marks,black,dashed,line width=1.2pt,domain=0.1:1.1,samples=100] {0.375*(sin(540*((x)-0.6)) + sin(180*((x)-0.6))) + 4.1};
\addplot+[no marks,black,dashed,line width=1.2pt,domain=1.1:2.1,samples=100] {0.625*(sin(540*((x)-0.6)) + sin(180*((x)-0.6))) + 4.1};
\draw [line width=1.2pt,dashed] (axis cs: 2.1,4.1) -- (axis cs: 2.8,4.1);
\draw [line width=1.2pt,red,dashed] (axis cs: 0,4.15) -- (axis cs: 0.3,4.15);
\addplot+[no marks,red,dashed,line width=1.2pt,domain=0.3:1.3,samples=100] {0.375*(sin(540*((x)-0.8)) + sin(180*((x)-0.8))) + 4.15};
\addplot+[no marks,red,dashed,line width=1.2pt,domain=1.3:2.3,samples=100] {0.625*(sin(540*((x)-0.8)) + sin(180*((x)-0.8))) + 4.15};
\draw [line width=1.2pt,red,dashed] (axis cs: 2.3,4.15) -- (axis cs: 2.8,4.15);
\draw (axis cs: 1.7959,-2.3) -- (axis cs: 1.9959,-2.3) node[pos=0.5, below] {$\ell$};
\draw (axis cs: 1.7959,-2.2) -- (axis cs: 1.7959,-2.4);
\draw (axis cs: 1.9959,-2.2) -- (axis cs: 1.9959,-2.4);
\draw (axis cs: 2.5,-1.9245) -- (axis cs: 2.5,1.9245) node[pos=0.7,right] {$h_0$};
\draw (axis cs: 2.45,-1.9245) -- (axis cs: 2.55,-1.9245);
\draw (axis cs: 2.45,1.9245) -- (axis cs: 2.55,1.9245);
\draw (axis cs: 2.5,3.1277) -- (axis cs: 2.5,5.0623) node[pos=0.8,right] {$h_\alpha$};
\draw (axis cs: 2.45,3.1277) -- (axis cs: 2.55,3.1277);
\draw (axis cs: 2.45,5.0623) -- (axis cs: 2.55,5.0623);
\end{axis}
\end{tikzpicture}
\caption{
$f$ (solid, black) and $g$ (solid, red) are translations of each other, the shifted and renormalised signals are $\hat{f}=\frac{f+\alpha}{\int (f + \alpha)}$ (dashed, black) and $\hat{g}=\frac{g+\alpha}{\int (g + \alpha)}$ (dashed, red).
$d_{\mathrm{OT}}(\hat{f},\hat{g}) = O\left( \frac{h_0\ell}{\alpha}\right)$ whilst $TL^p$ requires no renormalization.
}
\label{subfig:OTandTLp:Rev:TL2vOT:TL2ReNorm}
\end{subfigure}
\hfill
\begin{subfigure}{0.45\columnwidth}
\centering
\scriptsize
\begin{tikzpicture}
\begin{axis}[%
width=\figurewidth,
height=\figureheight,
scale only axis,
xmin=-0.06667,
xmax=1.06667,
ymin=-0.25,
ymax=3.75,
hide axis,
name=plot2,
axis background/.style={fill=white!100}
]
\draw [->] (axis cs: 0,0) -- (axis cs: 1,0);
\draw [->] (axis cs: 0,0) -- (axis cs: 0,3.5);
\addplot+[no marks,black,line width=1.2pt,domain=0:0.8,samples=200] {(cos((x)*360)/4 + (x)^(1.5))/((x)+1)+1.7};
\addplot+[no marks,red,line width=1.2pt,domain=0:0.8,samples=500] {(cos((x)*360)/4 + (x)^(1.5))/((x)+1)+1.7+0.7*sin(3600*(x))};
\draw (axis cs: 0.125,2.7) -- (axis cs: 0.225,2.7) node[pos=0.5, above] {$\omega$};
\draw (axis cs: 0.125,2.65) -- (axis cs: 0.125,2.75);
\draw (axis cs: 0.225,2.65) -- (axis cs: 0.225,2.75);
\draw (axis cs: 0.82,2.175) -- (axis cs: 0.82,2.875) node[pos=0.5, right] {$A$};
\draw (axis cs: 0.81,2.175) -- (axis cs: 0.83,2.175);
\draw (axis cs: 0.81,2.875) -- (axis cs: 0.83,2.875);
\end{axis}
\end{tikzpicture}
\caption{
$f$ (black) is a low frequency signal and $g$ is a high frequency signal (red).
$d_{\mathrm{OT}}(f,g)= O(\omega) \ll 1$ whilst $d_{TL^p}(f,g)=O(A)$.
\vspace{23pt}
}
\label{subfig:OTandTLp:Rev:TL2vOT:TL2HighFreq}
\end{subfigure}
\caption{
A Comparison of $TL^p$ with OT.
}
\label{fig:OTandTLp:Rev:TL2vOT}
\end{figbox}
\end{figure*}

\begin{figure*}[ht]
\begin{figbox}
\centering
\setlength\figureheight{3cm}
\setlength\figurewidth{0.4\columnwidth}
\scriptsize
\input{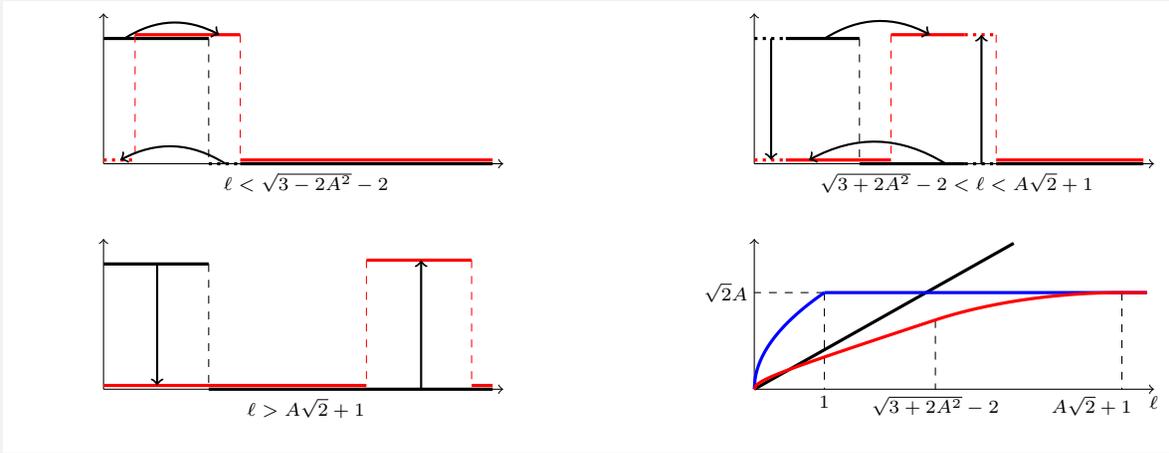}
\caption{
$TL^2$ transport between $f(x) = A\chi_{[0,1]}$ (black) and $g(x) = f(x-\ell)$ (red) and the $TL^2$ distance (red), $L^2$ distance (blue) and OT (black) between $f$ and $g$ (bottom right).
}
\label{fig:OTandTLp:Rev:Trans}
\end{figbox}
\end{figure*}

\paragraph*{Ability of $TL^p$ to Track Translations.}
Another desirable property of both $TL^p$ and OT are their ability to keep track of translations for further than $L^p$.
Let $f=A\chi_{[0,1]}$ be the indicator function of the set $[0,1]$ on $\mathbb{R}$ scaled by $A>1$ and $g(x)=f(x-\ell)$ the translation of $f$ by $\ell$.
Once $\ell>1$ then $L^p$ can no longer tell how far apart two humps are.
On the other hand OT can track the hump indefinitely.
In this example the $TL^p$ distance couples the graphs of $f$ and $g$ in one of three ways, see Figure~\ref{fig:OTandTLp:Rev:Trans}.
The first is when the transport is horizontal only in the graph (Figure~\ref{fig:OTandTLp:Rev:Trans} top left).
In the second (top right) there is a mixture of horizontal and vertical transport.
And in the third there is only vertical transport (bottom left), in which case the $TL^p$ distance coincides with the $L^p$ distance.
One can calculate the range of the $TL^p$ distance which is on the order of $A$.

\section{Definitions and Basic Properties of \texorpdfstring{$TL^p$}{TLp} \label{sec:Frame}}

In the previous section we defined the $TL^p$ distance for signals defined with respect to the Lebesgue measure.
In this section we generalise to signals defined on a general class of measures.
In particular we treat a signal as a pair $(f,\mu)$ where $f\in L^p(\mu;\mathbb{R}^m)$ for a measure $\mu\in \mathcal{P}_p(\Omega)$ (the set of probability measures with finite $p^{\text{th}}$ moment) and a function $f:\Omega\to \mathbb{R}^m$.
The general framework allows us to treat signal and discrete signals within the same framework as well as allowing one to design the underlying measure in order to emphasise certain parts of the signal.
We are also able to compare signals with different discretisations.
However, unless otherwise stated, $\mu=\nu$ is the Lebesgue measure.
In addition there is no assumption on the dimension $m$ of the codomain.
This allows us to consider multi-channelled signals.

The $TL_\lambda^p$ distance for pairs $(f,\mu) \in TL^p$ where
\[ TL^p := \left\{ (f,\mu) \, : \, f\in L^p(\mu), \mu \in\mathcal{P}_p(\Omega) \right\} \]
is defined by
\begin{align}
d_{TL^p_\lambda}^p((f,\mu),(g,\nu)) & = \min_{\pi \in \Pi(\mu,\nu)} \int_{\Omega\times\Omega} c_\lambda(x,y;f,g) \; \mathrm{d} \pi \label{eq:Frame:TLpd} \\
c_\lambda(x,y;f,g) & = \frac{1}{\lambda} |x-y|_p^p + |f(x)-g(y)|^p_p \label{eq:Frame:TLpc}
\end{align}
and $\Pi(\mu,\nu)$ is the space of measures on $\Omega\times\Omega$ such that the first marginal is $\mu$ and the second marginal is $\nu$.
Note that if $f=g$ is constant then we recover the OT distance between the measures $\mu$ and $\nu$.
In the special cases, when $\mu=\nu=\mathcal{L}$ are the Lebesgue measure, we write $d_{TL^p_\lambda}(f,g) := d_{TL^p_\lambda}((f,\mathcal{L}),(g,\mathcal{L}))$ and, when $\lambda=1$, $d_{TL^p}(f,g) := d_{TL^p_1}(f,g)$.
The result of~\cite[Proposition 3.3]{garciatrillos15} implies that $d_{TL_\lambda^p}$ is a metric on $TL^p$.

\begin{proposition}
\label{prop:Frame:Metric}
\cite{garciatrillos15}
For any $p\in [1,\infty]$, $(TL^p,d_{TL_\lambda^p})$ is a metric space.
\end{proposition}

When $\mu=\nu=\mathcal{L}$ is the Lebesgue measure then an admissible plan is the identity plan: $\pi(A\times B)=\mathcal{L}(A\cap B)$.
This implies that the $TL_\lambda^p$ distance is bounded above by the $L^p$ distance (for any $\lambda$).

In fact the parameter $\lambda$ controls how close the distance is to an $L^p$ distance.
As $\lambda\to 0$ then the cost of transport: $\frac{1}{\lambda}\int_{\Omega\times\Omega} |x-y|_p^p \; \mathrm{d} \pi(x,y)$, is very expensive which favours transport plans that are approximately the identity mapping.
Hence $d_{TL_0^p}(f,g) := \lim_{\lambda\to 0} d_{TL_\lambda^p}(f,g) = \|f-g\|_{L^p}$.
The following result, and the remainder of the results in this section, can be found in~\cite{thorpe16b}.

\begin{proposition}
\label{prop:Frame:lambdato0}
\cite{thorpe16b}
Let $f,g\in L^p$ (with respect to the Lebesgue measure).
The $TL^p_\lambda$ distance is decreasing as a function of $\lambda$ and
\[ \lim_{\lambda\to 0} d_{TL^p_\lambda}(f,g) = \| f-g\|_{L^p}. \]
Moreover, if either the derivative of $f$ or $g$ is bounded then
\[ d_{TL^p_\lambda}^p(f,g) \geq \left\{ \begin{array}{ll} \epsilon^{p-1}(\lambda) \| f-g\|^p_{L^p} & \text{if } p>1 \\ \| f-g\|^p_{L^p} & \text{if } p=1 \text{ and } \lambda < \frac{1}{\kappa} \end{array} \right.  \]
where $\epsilon(\lambda) = \frac{1}{1+(\lambda\kappa)^{\frac{1}{p-1}}}$ and $\kappa = \min\{\|D f\|_{L^\infty}^p,\|D g\|_{L^\infty}^p\}$.
\end{proposition}

The above proposition implies that, when $p=1$, if $\frac{1}{\lambda}$ is chosen larger than the length scale given by the derivative then the $TL_\lambda^1$ distance is exactly the $L^1$ distance.

Recall that we can consider the $TL^p_\lambda$ distance as an OT distance on the graphs of $f$ and $g$.
When there exists a map realising the minimum in $d_{TL_\lambda^p}$ then we can understand the transport as a map $(x,f(x))\mapsto (y,g(y))$.
We refer to the transport $x\mapsto y$ in the domain $\Omega$ as horizontal transport and transport $f(x)\mapsto g(y)$ in the codomain of $f$ and $g$ as vertical transport.
We see that horizontal transport is favoured as $\lambda\to \infty$.
For example, if we consider $f(x)=\chi_{[0,1]}$ and $g(x) = \chi_{[1,2]}$ defined on the interval $[0,2]$ then the mapping $T(x) = x+1$ if $x\in [0,1]$ and $T(x) = x-1$ otherwise has cost
\begin{align*}
d_{TL^p_\lambda}^p(f,g) & \leq \int_0^2 \frac{|x-T(x)|^p}{\lambda} + |f(x) - g(T(x))|^p \; \mathrm{d} x \\
 &= \frac{2}{\lambda} \to 0 \quad \text{as } \lambda \to \infty.
\end{align*}
In this example $d_{TL_\infty^p}(f,g) := \lim_{\lambda \to \infty} d_{TL_\lambda^p}(f,g) = 0$.
The $TL_\infty^p$ distance is an OT distance between the measures $f_{\#}\mu$ and $g_{\#}\nu$.

\begin{proposition}
\label{prop:Frame:lambdatoInfty}
\cite{thorpe16b}
Let $\Omega\subseteq\mathbb{R}^d$, $f,g:\Omega\to \mathbb{R}^m$ measurable functions and $\mu,\nu\in\mathcal{P}_p(\Omega)$ where $p\geq 1$, then
\[ \lim_{\lambda\to \infty} d_{TL_\lambda^p}((f,\mu),(g,\nu)) = d_{\mathrm{OT}}(f_{\#}\mu,g_{\#}\nu) \]
where $d_{\mathrm{OT}}$ is the OT distance (on $\mathcal{P}(\mathbb{R}^m)$) with cost $c(x,y) = |x-y|_p^p$.
\end{proposition}

As the example before the proposition showed, $d_{TL_\infty^p}(f,g)$ is not a metric, however is non-negative, symmetric and the triangle inequality holds.

We observe that when $\mu$ is a uniform measure (either in the discrete or continuous sense) the measure $f_{\#}\mu$ is the histogram of $f$.
The OT distance between histograms is a popular tool in histogram specification.
Minimizers to the Monge formulation of $d_{\mathrm{OT}}(f_{\#}\mu,g_{\#}\nu)$ define a mapping between the histograms $f_{\#}\mu$ and $g_{\#}\nu$~\cite{morovic03,rabin14,rabin15}.
However this mapping contains no spatial information.
If instead one uses minimizers to the Monge formulation of the $TL^p_\lambda$ distance~\eqref{eq:Frame:TLpMonge} ($\lambda<\infty$) then one can include spatial information in the histogram specification.
We explore this further in Section~\ref{subsec:App:Colour} and apply the method to the colour transfer problem.

It is well known that there exists a minimizer (when $c$ is lower semi-continuous) for OT.
Since $d_{TL_\lambda^p}$ is closely related to an OT distance between measures in $\mathbb{R}^{d+m}$ (i.e. measures supported on graphs) then there exists a minimizer to $TL^p_\lambda$.

\begin{proposition}
\label{prop:Frame:ExistTransPlans}
\cite{thorpe16b}
Let $\Omega\subseteq \mathbb{R}^d$ be open and bounded, $f\in L^p(\mu)$, $g\in L^p(\nu)$ where $\mu,\nu\in\mathcal{P}(\Omega)$, $\lambda\in [0,+\infty]$ and $p\geq 1$.
Under these conditions there exists an optimal plan $\pi\in \Pi(\mu,\nu)$ realising the minimum in $d_{TL^p_\lambda}((f,\mu),(g,\nu))$.
\end{proposition}

As in the OT case it is natural to set the $TL_\lambda^p$ problem in the Monge formulation~\eqref{eq:OTandTLp:Rev:MOT}
\begin{equation} \label{eq:Frame:TLpMonge}
d_{TL_\lambda^{p}}((f,\mu),(g,\nu)) = \inf_{T : T_{\#}\mu=\nu} \int_{\Omega} c_\lambda(x,T(x);f,g) \; \mathrm{d} \mu(x).
\end{equation}
Minimizers to the above will not always exist.
For example, consider when $f=g$ then the $TL_\lambda^p$ distance is the OT distance between $\mu$ and $\nu$.
If one chooses $\mu=\frac{1}{3}\delta_{x_1}+\frac{1}{3}\delta_{x_2}+\frac{1}{3}\delta_{x_3}$ and $\nu=\frac{1}{2}\delta_{y_1}+\frac{1}{2}\delta_{y_2}$ where all of $x_i,y_j$ are distinct then there are no maps $T:\{x_1,x_2,x_3\}\to \{ y_1,y_2\}$ that pushforward $\mu$ to $\nu$.

However, in terms of numerical implementation, an interesting and important case is when $\mu$ and $\nu$ are discrete measures (see also~\cite[pg 5, 14-15]{villani03} for the following argument with the Monge OT problem).
Let $\mu=\frac{1}{n}\sum_{i=1}^n \delta_{x_i}$ and $\nu = \frac{1}{n} \sum_{i=1}^n \delta_{y_i}$ then $\pi = (\pi_{ij})_{i,j=1}^n\in \Pi(\mu,\nu)$ is a doubly stochastic matrix up to a factor of $\frac{1}{n}$, that is
\begin{equation} \label{eq:Frame:Def:DoubStoch}
\pi_{ij} \geq 0 \, \forall i, j, \, \, \sum_{i=1}^n \pi_{ij} = \frac{1}{n} \, \forall j \, \, \text{and} \, \, \sum_{j=1}^n \pi_{ij} = \frac{1}{n} \, \forall i,
\end{equation}
and the $TL^{p}_\lambda$ distance can be written
\begin{equation} \label{eq:Frame:Def:LP}
d_{TL_\lambda^{p}}^p((f,\mu),(g,\nu)) = \min \sum_{i=1}^n \sum_{j=1}^n c_\lambda(x_i,y_j;f,g) \pi_{ij}
\end{equation}
where the minimum is taken over $\pi$ satisfying~\eqref{eq:Frame:Def:DoubStoch}.
It is known (by Choquet's Theorem) that the solution to this minimisation problem is an extremal point in the matrix set $\Pi(\mu,\nu)$.
It is also known (by Birkhoff's Theorem) that extremal points in $\Pi(\mu,\nu)$ are permutation matrices.
This implies that there exists an optimal plan $\pi^*$ that can be written as $\pi^*_{ij} = \frac{1}{n}\delta_{j-\sigma(i)}$ for a permutation $\sigma:\{1,\dots,n\}\to\{1,\dots,n\}$.
Hence there exists an optimal plan to the Monge formulation of $TL_\lambda^{p}$.

\begin{proposition}
\label{prop:Frame:Def:ExistTransMap}
For any $f\in L^p(\mu)$ and $g\in L^p(\nu)$ where $\mu = \frac{1}{n}\sum_{i=1}^n \delta_{x_i}$ and $\nu = \frac{1}{n} \sum_{j=1}^n \delta_{y_j}$ there exists a permutation $\sigma:\{1,2,\dots,n\} \to \{ 1,2,\dots,n\}$ such that
\[ d_{TL_\lambda^p}((f,\mu),(g,\nu)) = \frac{1}{n} \sum_{i=1}^n c_\lambda(x_i,x_{\sigma(i)};f,g). \]
\end{proposition}

The above theorem implies that in the discrete case there exists optimal plans (which are matrices) which will be sparse.
In particular, $\pi^*$ is an $n\times n$ matrix with only $n$ non-zero entries.
This motivates the use of numerical methods that can take advantage of expected sparsity in the solution (e.g. iterative linear programming methods such as~\cite{oberman15}).

\section{\texorpdfstring{$TL^p$}{TLp} in Multivariate Signal and Image Processing \label{sec:App}}

Written in the form~\eqref{eq:Frame:TLpd} the $TL^p$ distance is an optimal transport problem between the measures $\mu$ and $\nu$ with the cost function $c$ given by~\eqref{eq:Frame:TLpc} and which depends upon $f$ and $g$.
Hence, to compute $TL^p$ there are many algorithms for OT that we may apply, for example the multi-scale approaches of Schmitzer~\cite{schmitzer16} and Oberman and Ruan~\cite{oberman15}, or the entropy regularized approaches of Cuturi~\cite{cuturi13} and Benamou, Carlier, Cuturi, Nenna and Peyre~\cite{benamou15}.
Our choice was the iterative linear programming method of Oberman and Ruan~\cite{oberman15} for the multivariate signals which we find works well both in terms of accuracy and computation time.
Our choice for the images was the entropy regularized solution due to Cuturi~\cite{benamou15,cuturi13}.
Whilst this only produces an approximation of the $TL^p$ distance we find it computationally efficient for 2D images.
For convenience we include a review of the numerical methods in Appendix~\ref{sec:app:Numerics}.

With respect to choosing $\lambda$ there are two approaches we could take.
The first is to compute the $TL^p$ distance for a range of $\lambda$ and then use cross-validation.
There are two disadvantages to this approach: we would still have to know the range of $\lambda$ and computing the $TL_\lambda^p$ distance for multiple choices of $\lambda$ would considerably increase computation time.
The second approach, and the one we use for each example in this section, is to estimate $\lambda$ by comparing length scales and desired behaviour.
In particular we choose $\lambda$ so that both horizontal and vertical transport make a contribution.
For the applications in this section we want to stay away from the asymptotic regimes $\lambda\approx 0$ and $\lambda\gg 1$.
By balancing the vertical and horizontal length scale we can formally find an approximation of $\lambda$ which in our results below works well.

We first consider two synthetic examples.
Considering synthetic examples allows us to better demonstrate where $TL^p$ will be successful.
In particular synthetic examples can simplify the analysis and allow us to draw attention to features that may be obscured in real world applications.

The first synthetic example considers three classes where we can analytically compute the within class distances and between class separation.
This allows us to compare how well we expect $TL^p$ to perform in a classification problem.

The second synthetic example uses simulated 2D data from one-hump and two-hump functions.
We test how well $TL^p$ recovers the classes and compare with OT and $L^p$.

Our first real world application is to classifying multivariate times series and 2D images.
We choose a multivariate time series data set where we expect transport based methods to be successful but it is not clear how one could apply OT distances (one would want to define a `multi-valued measure').
Our chosen data set consists of sequences of sign language data (we define the data set in more detail shortly) which contains the position of both hands (parametrised by 22 variables) at each time.
The $TL^p$ distance can treat these signals as functions $f:[0,1]\to \mathbb{R}^{22}$.
We expect to see certain features in the signals however these may be shifted based on the speed of the speaker.
The second data set contains 2D images that must be normalised in order to apply the OT distance, this distorts some of the features leading to a poor performance.

The second real world application is to histogram specification and colour transfer.
Histogram specification or matching, where one defines a map $T$ that matches one histogram with another, is widely used to define a colour transfer scheme.
In particular let $f:\{x_i\}_{i=1}^N\to \mathbb{R}^3$ represent a colour image by mapping pixels $x_i$ to a colour $f(x_i)$ (for example in RGB space), one defines a multidimensional histogram of colours on an image by $\varphi(c) = \frac{1}{N}\#\{ x_i \, : \, f(x_i) = c\}$.
For colour images the histogram $\varphi$ is a measure on $\mathbb{R}^3$.
For notational clarity we will call $\varphi$ the colour histogram.
One can equivalently define a histogram for grayscale images as a measure on $\mathbb{R}$.

Let $\varphi$ and $\psi$ be two colour histograms for images $f$ and $g$ respectively.
The OT map $T$ defines a rearrangement of $\varphi$ onto $\psi$, that is $\psi = T_{\#} \varphi$.
In colour transfer the map $T$ is used to colour the image $f$ using the palette of $g$ by $\hat{f}(x) = g(T(x))$.

The histogram contains only intensity information and in particular there is no spatial dependence.
Using the $TL_\lambda^p$-optimal map we define a spatially correlated histogram specification and explain how this can be applied to the colour transfer problem.

\subsection{1D Class Separation for Synthetic Data \label{subsec:App:Syn1}}

\paragraph*{Objective.}
We compare the expected classification power of $TL^p$, $L^p$ and OT with three classes of 1D signals that differ by position (translations), shape (1 hump versus 2 hump) and frequency (hump versus chirp).

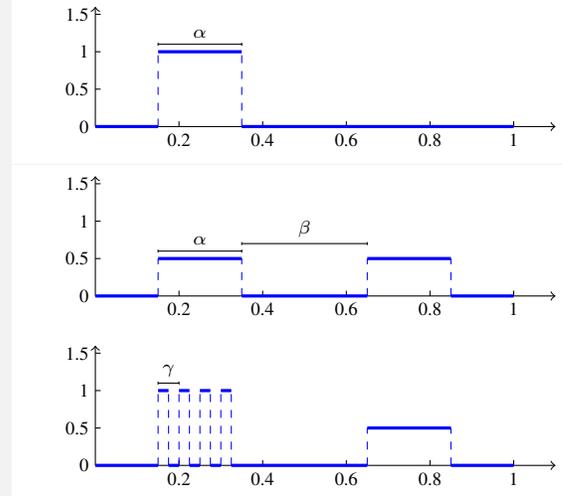
\begin{figure*}
\begin{figbox}
\centering
\setlength\figureheight{2.8cm}
\setlength\figurewidth{\columnwidth}
\scriptsize
\begin{tikzpicture}

\begin{axis}[%
width=0.45\figurewidth,
height=0.8\figureheight,
scale only axis,
xmin=-0.2,
xmax=1.2,
ymin=-0.5,
ymax=1.75,
hide axis,
name=plot2,
axis background/.style={fill=white!100},
]
\draw [->] (axis cs: 0,0) -- (axis cs: 1.1,0);
\draw [->] (axis cs: 0,0) -- (axis cs: 0,1.6);
\draw[-] (axis cs: 0,0) -- ++(2pt,0) node[left,xshift=-2pt] {0};
\draw[-] (axis cs: 0,0.5) -- ++(2pt,0) node[left,xshift=-2pt] {0.5};
\draw[-] (axis cs: 0,1) -- ++(2pt,0) node[left,xshift=-2pt] {1};
\draw[-] (axis cs: 0,1.5) -- ++(2pt,0) node[left,xshift=-2pt] {1.5};
\draw[-] (axis cs: 0.2,0) -- ++(0,2pt) node[below,yshift=-2pt] {0.2};
\draw[-] (axis cs: 0.4,0) -- ++(0,2pt) node[below,yshift=-2pt] {0.4};
\draw[-] (axis cs: 0.6,0) -- ++(0,2pt) node[below,yshift=-2pt] {0.6};
\draw[-] (axis cs: 0.8,0) -- ++(0,2pt) node[below,yshift=-2pt] {0.8};
\draw[-] (axis cs: 1,0) -- ++(0,2pt) node[below,yshift=-2pt] {1};
\draw (axis cs: 0.15,1.1) -- (axis cs: 0.35,1.1) node[pos=0.5, above] {$\alpha$};
\draw (axis cs: 0.15,1.08) -- (axis cs: 0.15,1.12);
\draw (axis cs: 0.35,1.08) -- (axis cs: 0.35,1.12);
\draw[blue,line width=1.2pt] (axis cs: 0,0) -- (axis cs: 0.15,0);
\draw[blue,line width=1.2pt] (axis cs: 0.15,1) -- (axis cs: 0.35,1);
\draw[blue,line width=1.2pt] (axis cs: 0.35,0) -- (axis cs: 1,0);
\draw[blue,dashed] (axis cs: 0.15,0) -- (axis cs: 0.15,1);
\draw[blue,dashed] (axis cs: 0.35,0) -- (axis cs: 0.35,1);
\end{axis}

\begin{axis}[%
width=0.55\figurewidth,
height=\figureheight,
scale only axis,
xmin=-0.2,
xmax=1.2,
ymin=-0.5,
ymax=1.75,
hide axis,
name=plot1,
at=(plot2.left of south west),
anchor=right of south east
]
\node[text width=0.55\figurewidth] at (axis cs: 0.5,0.625) {\normalsize ($\mathcal{C}_1$) \textbf{One hump functions:} of the form \[ f = \chi_{[\ell,\ell+\alpha]} \] where $\ell \in [0,1-\alpha]$.};
\end{axis}

\begin{axis}[%
width=0.45\figurewidth,
height=0.8\figureheight,
scale only axis,
xmin=-0.2,
xmax=1.2,
ymin=-0.5,
ymax=1.75,
hide axis,
name=plot4,
axis background/.style={fill=white!100},
at=(plot2.below south west),
anchor=above north west,
]
\draw [->] (axis cs: 0,0) -- (axis cs: 1.1,0);
\draw [->] (axis cs: 0,0) -- (axis cs: 0,1.6);
\draw[-] (axis cs: 0,0) -- ++(2pt,0) node[left,xshift=-2pt] {0};
\draw[-] (axis cs: 0,0.5) -- ++(2pt,0) node[left,xshift=-2pt] {0.5};
\draw[-] (axis cs: 0,1) -- ++(2pt,0) node[left,xshift=-2pt] {1};
\draw[-] (axis cs: 0,1.5) -- ++(2pt,0) node[left,xshift=-2pt] {1.5};
\draw[-] (axis cs: 0.2,0) -- ++(0,2pt) node[below,yshift=-2pt] {0.2};
\draw[-] (axis cs: 0.4,0) -- ++(0,2pt) node[below,yshift=-2pt] {0.4};
\draw[-] (axis cs: 0.6,0) -- ++(0,2pt) node[below,yshift=-2pt] {0.6};
\draw[-] (axis cs: 0.8,0) -- ++(0,2pt) node[below,yshift=-2pt] {0.8};
\draw[-] (axis cs: 1,0) -- ++(0,2pt) node[below,yshift=-2pt] {1};
\draw (axis cs: 0.15,0.6) -- (axis cs: 0.35,0.6) node[pos=0.5, above] {$\alpha$};
\draw (axis cs: 0.15,0.58) -- (axis cs: 0.15,0.62);
\draw (axis cs: 0.35,0.58) -- (axis cs: 0.35,0.62);
\draw (axis cs: 0.35,0.7) -- (axis cs: 0.65,0.7) node[pos=0.5, above] {$\beta$};
\draw (axis cs: 0.35,0.68) -- (axis cs: 0.35,0.72);
\draw (axis cs: 0.65,0.68) -- (axis cs: 0.65,0.72);
\draw[blue,line width=1.2pt] (axis cs: 0,0) -- (axis cs: 0.15,0);
\draw[blue,line width=1.2pt] (axis cs: 0.15,0.5) -- (axis cs: 0.35,0.5);
\draw[blue,line width=1.2pt] (axis cs: 0.35,0) -- (axis cs: 0.65,0);
\draw[blue,line width=1.2pt] (axis cs: 0.65,0.5) -- (axis cs: 0.85,0.5);
\draw[blue,line width=1.2pt] (axis cs: 0.85,0) -- (axis cs: 1,0);
\draw[blue,dashed] (axis cs: 0.15,0) -- (axis cs: 0.15,0.5);
\draw[blue,dashed] (axis cs: 0.35,0) -- (axis cs: 0.35,0.5);
\draw[blue,dashed] (axis cs: 0.65,0) -- (axis cs: 0.65,0.5);
\draw[blue,dashed] (axis cs: 0.85,0) -- (axis cs: 0.85,0.5);
\end{axis}

\begin{axis}[%
width=0.55\figurewidth,
height=\figureheight,
scale only axis,
xmin=-0.2,
xmax=1.2,
ymin=-0.5,
ymax=1.75,
hide axis,
name=plot3,
at=(plot4.left of south west),
anchor=right of south east
]
\node[text width=0.55\figurewidth] at (axis cs: 0.5,0.625) {\normalsize ($\mathcal{C}_2$) \textbf{Two hump functions:} of the form \[ f=\frac{1}{2}\left(\chi_{[\ell,\ell+\alpha]} + \chi_{[\ell+\beta+\alpha,\ell+\beta+2\alpha]}\right) \] where $\ell \in [0,1-\beta-2\alpha]$.};
\end{axis}

\begin{axis}[%
width=0.45\figurewidth,
height=0.8\figureheight,
scale only axis,
xmin=-0.2,
xmax=1.2,
ymin=-0.5,
ymax=1.75,
hide axis,
at=(plot4.below south west),
anchor=above north west,
name=plot6,
axis background/.style={fill=white!100},
]
\draw [->] (axis cs: 0,0) -- (axis cs: 1.1,0);
\draw [->] (axis cs: 0,0) -- (axis cs: 0,1.6);
\draw[-] (axis cs: 0,0) -- ++(2pt,0) node[left,xshift=-2pt] {0};
\draw[-] (axis cs: 0,0.5) -- ++(2pt,0) node[left,xshift=-2pt] {0.5};
\draw[-] (axis cs: 0,1) -- ++(2pt,0) node[left,xshift=-2pt] {1};
\draw[-] (axis cs: 0,1.5) -- ++(2pt,0) node[left,xshift=-2pt] {1.5};
\draw[-] (axis cs: 0.2,0) -- ++(0,2pt) node[below,yshift=-2pt] {0.2};
\draw[-] (axis cs: 0.4,0) -- ++(0,2pt) node[below,yshift=-2pt] {0.4};
\draw[-] (axis cs: 0.6,0) -- ++(0,2pt) node[below,yshift=-2pt] {0.6};
\draw[-] (axis cs: 0.8,0) -- ++(0,2pt) node[below,yshift=-2pt] {0.8};
\draw[-] (axis cs: 1,0) -- ++(0,2pt) node[below,yshift=-2pt] {1};
\draw (axis cs: 0.15,1.1) -- (axis cs: 0.2,1.1) node[pos=0.5, above] {$\gamma$};
\draw (axis cs: 0.15,1.08) -- (axis cs: 0.15,1.12);
\draw (axis cs: 0.2,1.08) -- (axis cs: 0.2,1.12);
\draw[blue,line width=1.2pt] (axis cs: 0,0) -- (axis cs: 0.15,0);
\draw[blue,line width=1.2pt] (axis cs: 0.15,1) -- (axis cs: 0.175,1);
\draw[blue,line width=1.2pt] (axis cs: 0.2,1) -- (axis cs: 0.225,1);
\draw[blue,line width=1.2pt] (axis cs: 0.25,1) -- (axis cs: 0.275,1);
\draw[blue,line width=1.2pt] (axis cs: 0.3,1) -- (axis cs: 0.325,1);
\draw[blue,line width=1.2pt] (axis cs: 0.175,0) -- (axis cs: 0.2,0);
\draw[blue,line width=1.2pt] (axis cs: 0.225,0) -- (axis cs: 0.25,0);
\draw[blue,line width=1.2pt] (axis cs: 0.275,0) -- (axis cs: 0.3,0);
\draw[blue,line width=1.2pt] (axis cs: 0.325,0) -- (axis cs: 0.65,0);
\draw[blue,line width=1.2pt] (axis cs: 0.65,0.5) -- (axis cs: 0.85,0.5);
\draw[blue,line width=1.2pt] (axis cs: 0.85,0) -- (axis cs: 1,0);
\draw[blue,dashed] (axis cs: 0.15,0) -- (axis cs: 0.15,1);
\draw[blue,dashed] (axis cs: 0.175,0) -- (axis cs: 0.175,1);
\draw[blue,dashed] (axis cs: 0.2,0) -- (axis cs: 0.2,1);
\draw[blue,dashed] (axis cs: 0.225,0) -- (axis cs: 0.225,1);
\draw[blue,dashed] (axis cs: 0.25,0) -- (axis cs: 0.25,1);
\draw[blue,dashed] (axis cs: 0.275,0) -- (axis cs: 0.275,1);
\draw[blue,dashed] (axis cs: 0.3,0) -- (axis cs: 0.3,1);
\draw[blue,dashed] (axis cs: 0.325,0) -- (axis cs: 0.325,1);
\draw[blue,dashed] (axis cs: 0.65,0) -- (axis cs: 0.65,0.5);
\draw[blue,dashed] (axis cs: 0.85,0) -- (axis cs: 0.85,0.5);
\end{axis}

\begin{axis}[%
width=0.55\figurewidth,
height=\figureheight,
scale only axis,
xmin=-0.2,
xmax=1.2,
ymin=-0.5,
ymax=1.75,
hide axis,
name=plot3,
at=(plot6.left of north west),
anchor=right of north east
]
\node[text width=0.55\figurewidth] at (axis cs: 0.5,0.625) {\normalsize ($\mathcal{C}_3$) \textbf{One hump, one chirp functions:} of the form \[ f = \sum_{i=0}^{\frac{\alpha}{\gamma}-1} \chi_{\left[\ell+i\gamma,\ell+\frac{(2i+1)\gamma}{2}\right]} + \frac{1}{2}\chi_{[\ell+\beta+\alpha,\ell+\beta+2\alpha]} \] where $\ell \in [0,1-\beta-2\alpha]$.};
\end{axis}

\end{tikzpicture}
\caption{
For fixed $\alpha,\beta,\gamma \in (0,1)$ where $\beta>\alpha\gg \gamma$ the definition of the classes $\mathcal{C}_i$.
}
\label{fig:App:Syn1:Classes}
\end{figbox}
\end{figure*}

\paragraph*{Data Sets.}
We consider data from three classes defined in Figure~\ref{fig:App:Syn1:Classes}.
The first class contains single hump function and the second class contains two hump functions.
The third class consists of functions with one hump and one chirp, defined to be a high frequency perturbation of a hump.
The classes are chosen to test the performance of $TL^2$ with $L^2$ and OT with regards to identifying translations (where we expect $L^2$ to do poorly) with a class containing high frequency perturbations (where we expect OT to do poorly).

\paragraph*{Methods.}
For a distance to have good performance in classification and clustering problems it should be able to separate classes.
To be able to quantify this we use the ratio of `between class separation' to `class coverage radius' that we define now.

Let $\mathcal{C}_i^N = \{f_j^i\}_{j=1}^N$ be a sample of $N$ functions from class $\mathcal{C}_i$.
For a given radius $r$ we let $G_i(r)$ be the graph defined by connecting any two points in $\mathcal{C}_i^N$ with distance less than $r$.
The distance will be defined using the $TL^p_\lambda$, $L^2$ and OT metrics.
Let $R_{TL^p_\lambda}(\mathcal{C}_i^N)$ be the smallest $r$ such that $G_i(r)$ is a connected graph using the $TL^p_\lambda$ metric.
Analogously we can define $R_{L^p}$ and $R_{\mathrm{OT}}$.

We define `between class separation' as the Hausdorff distance between classes:
\begin{align*}
& d_{H,\rho}(\mathcal{C}_i^N,\mathcal{C}_j^N) = \\
& \max\left\{ \sup_{f\in \mathcal{C}_i^N} \inf_{g\in \mathcal{C}_j^N} \rho(f,g), \sup_{g\in \mathcal{C}_j^N} \inf_{f\in \mathcal{C}_i^N} \rho(f,g) \right\}
\end{align*}
where we will consider $\rho$ to be one of the $TL^2$, $L^2$ or OT metrics.
Large values of $d_{H,\rho}(\mathcal{C}_i^N,\mathcal{C}_j^N)$ imply that the classes $\mathcal{C}_i^N$ and $\mathcal{C}_j^N$ are well separated.

When $R_\rho(\mathcal{C}_i^N) \leq d_{H,\rho}(\mathcal{C}_i^N,\mathcal{C}_j^N)$ then we say that the class $\mathcal{C}_i^N$ is separable from class $\mathcal{C}_j^N$ since for any $f\in \mathcal{C}_i^N$ the nearest neighbour in $\left(\mathcal{C}_i^N\cup\mathcal{C}_j^N\right)\setminus\{f\}$ is also in class $\mathcal{C}_i^N$.
We define the pairwise property
\[ \kappa_{ij}(\rho;N) = \frac{\mathbb{E}d_{H,\rho}(\mathcal{C}_i^N,\mathcal{C}_j^N)}{\max\{\mathbb{E}R_\rho(\mathcal{C}_i^N),\mathbb{E}R_\rho(\mathcal{C}_j^N)\}} \]
where we take the expectation over sample classes $\mathcal{C}_i^N$.
We will assume that the distribution over each class is uniform in the parameter $\ell$.
When $\kappa_{ij}(\rho;N)>1$ then we expect classes $\mathcal{C}_i^N$ and $\mathcal{C}_j^N$ to be separable from each other.

As a performance metric we use the smallest value of $N$ such that $\kappa_{ij}(\rho;N)\geq 1$.
We let
\[ N^*_{ij}(\rho) = \min \{N \, : \, \kappa_{ij}(\rho;N)\geq 1\}. \]  
This measures how many data points we need in order to expect a good classification accuracy.

\paragraph*{Results.}
We leave the calculation to the appendix but the conclusion is 
\begin{align*}
& N^*_{12}(TL^2) < N^*_{12}(\mathrm{OT}) < N^*_{12}(L^2) \\
& N^*_{13}(TL^2) < N^*_{13}(\mathrm{OT}) < N^*_{13}(L^2) \\
& N^*_{23}(TL^2) < N^*_{23}(\mathrm{OT}) < N^*_{23}(L^2).
\end{align*}
In each case the $TL^2$ distance outperforms $L^2$ and OT.

In each class the $L^2$ distance has a larger value of $R$.
This implies a larger data set is needed to accurately cover each class.
This is due to the Lagrangian nature of signals within each class (translations) that is poorly represented by $L^2$.
OT has the lowest (and therefore best) value of $R$ in each class.
Since each class is Lagrangian then the OT distance is very small between functions of the same class.

When considering between class separation the $TL^2$ and $L^2$ distances coincide and give a bigger (and better) between class distance than OT.
Since the class $\mathcal{C}_3$ can be written as a high frequency perturbation of functions in the class $\mathcal{C}_2$ then the OT distance struggles to tell the difference between these classes.
The distance $d_{H,\mathrm{OT}}(\mathcal{C}_2^N,\mathcal{C}_3^N)$ is comparatively small so that one needs more data points in order to fully resolve these classes.
We see a similar effect when considering $d_{H,\mathrm{OT}}$ for the other classes.

\subsection{2D Classification for Synthetic Data \label{subsec:App:Syn2}}

\paragraph*{Objective.}
We use simulated data to illustrate better separation of $TL^p$ compared to $L^p$ and OT distances for 2D data from two classes of 1-hump and 2-hump functions.

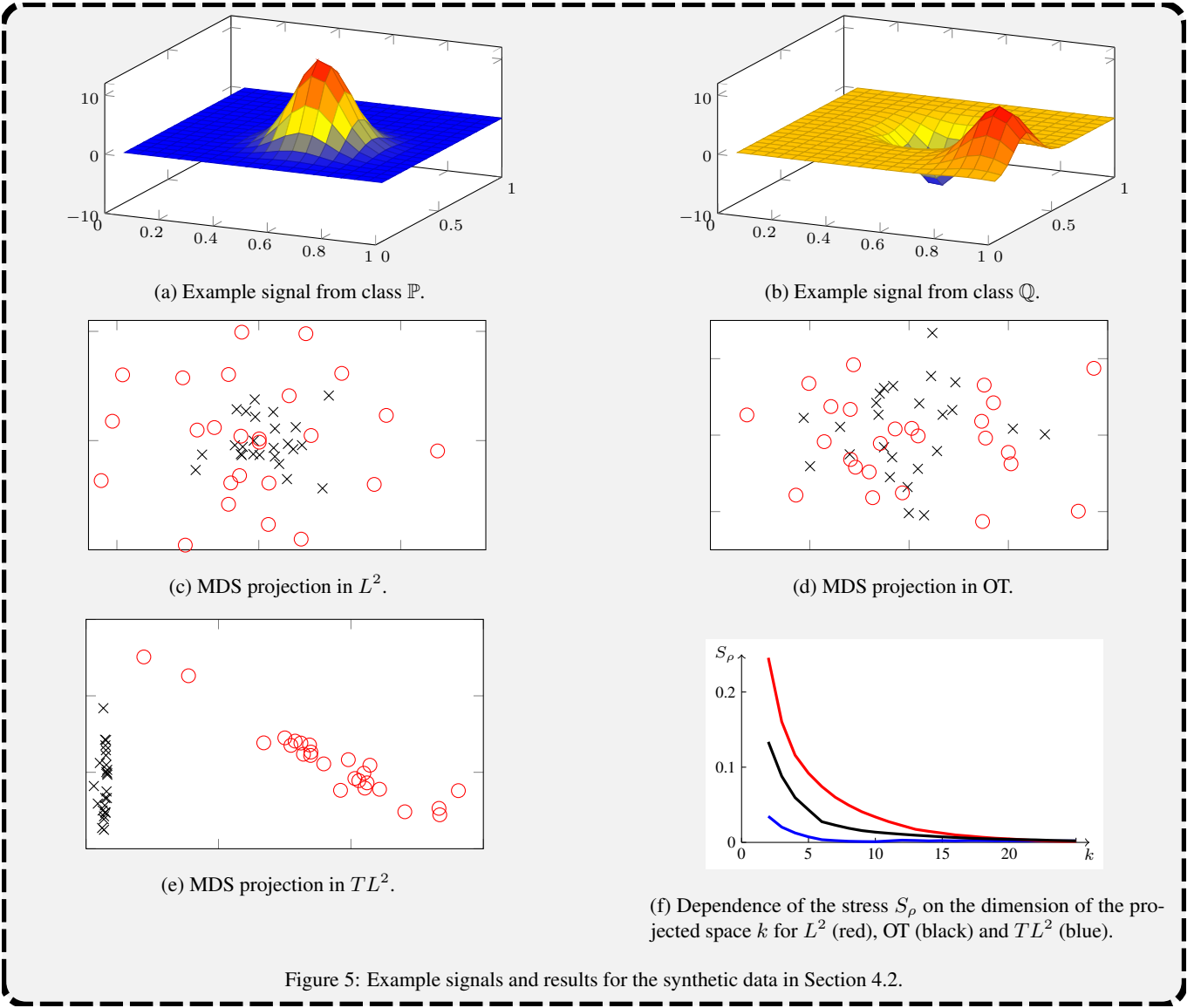
\begin{figure*}[ht]
\begin{figbox}
\centering
\setlength\figureheight{3.5cm}
\setlength\figurewidth{0.35\columnwidth}
\scriptsize
\begin{subfigure}[t]{0.45\columnwidth}
\centering
\scriptsize
\begin{tikzpicture} 
\begin{axis}[
width=\figurewidth,
height=\figureheight,
scale only axis,
xmin=-0,
xmax=1,
ymin=0,
ymax=1,
zmin=-10,
zmax=12,
axis background/.style={fill=white!100},
]
\addplot3[surf] coordinates{
(0.05,0.05,0.0000) (0.1,0.05,0.0000) (0.15,0.05,0.0000) (0.2,0.05,0.0000) (0.25,0.05,0.0000) (0.3,0.05,0.0001) (0.35,0.05,0.0003) (0.4,0.05,0.0011) (0.45,0.05,0.0032) (0.5,0.05,0.0070) (0.55,0.05,0.0114) (0.6,0.05,0.0141) (0.65,0.05,0.0132) (0.7,0.05,0.0094) (0.75,0.05,0.0051) (0.8,0.05,0.0021) (0.85,0.05,0.0006) (0.9,0.05,0.0002) (0.95,0.05,0.0000) (1,0.05,0.0000)  

(0.05,0.1,0.0000) (0.1,0.1,0.0000) (0.15,0.1,0.0000) (0.2,0.1,0.0000) (0.25,0.1,0.0001) (0.3,0.1,0.0004) (0.35,0.1,0.0019) (0.4,0.1,0.0071) (0.45,0.1,0.0201) (0.5,0.1,0.0432) (0.55,0.1,0.0705) (0.6,0.1,0.0873) (0.65,0.1,0.0819) (0.7,0.1,0.0583) (0.75,0.1,0.0314) (0.8,0.1,0.0128) (0.85,0.1,0.0040) (0.9,0.1,0.0009) (0.95,0.1,0.0002) (1,0.1,0.0000)  

(0.05,0.15,0.0000) (0.1,0.15,0.0000) (0.15,0.15,0.0000) (0.2,0.15,0.0000) (0.25,0.15,0.0003) (0.3,0.15,0.0018) (0.35,0.15,0.0088) (0.4,0.15,0.0331) (0.45,0.15,0.0940) (0.5,0.15,0.2025) (0.55,0.15,0.3307) (0.6,0.15,0.4094) (0.65,0.15,0.3842) (0.7,0.15,0.2733) (0.75,0.15,0.1474) (0.8,0.15,0.0602) (0.85,0.15,0.0187) (0.9,0.15,0.0044) (0.95,0.15,0.0008) (1,0.15,0.0001) 
 
(0.05,0.2,0.0000) (0.1,0.2,0.0000) (0.15,0.2,0.0000) (0.2,0.2,0.0001) (0.25,0.2,0.0010) (0.3,0.2,0.0063) (0.35,0.2,0.0314) (0.4,0.2,0.1176) (0.45,0.2,0.3342) (0.5,0.2,0.7199) (0.55,0.2,1.1756) (0.6,0.2,1.4553) (0.65,0.2,1.3656) (0.7,0.2,0.9715) (0.75,0.2,0.5239) (0.8,0.2,0.2141) (0.85,0.2,0.0664) (0.9,0.2,0.0156) (0.95,0.2,0.0028) (1,0.2,0.0004)  

(0.05,0.25,0.0000) (0.1,0.25,0.0000) (0.15,0.25,0.0000) (0.2,0.25,0.0003) (0.25,0.25,0.0026) (0.3,0.25,0.0171) (0.35,0.25,0.0845) (0.4,0.25,0.3168) (0.45,0.25,0.9004) (0.5,0.25,1.9396) (0.55,0.25,3.1675) (0.6,0.25,3.9211) (0.65,0.25,3.6796) (0.7,0.25,2.6175) (0.75,0.25,1.4115) (0.8,0.25,0.5770) (0.85,0.25,0.1788) (0.9,0.25,0.0420) (0.95,0.25,0.0075) (1,0.25,0.0010) 
 
(0.05,0.3,0.0000) (0.1,0.3,0.0000) (0.15,0.3,0.0001) (0.2,0.3,0.0006) (0.25,0.3,0.0053) (0.3,0.3,0.0349) (0.35,0.3,0.1726) (0.4,0.3,0.6471) (0.45,0.3,1.8390) (0.5,0.3,3.9617) (0.55,0.3,6.4695) (0.6,0.3,8.0088) (0.65,0.3,7.5155) (0.7,0.3,5.3462) (0.75,0.3,2.8829) (0.8,0.3,1.1785) (0.85,0.3,0.3652) (0.9,0.3,0.0858) (0.95,0.3,0.0153) (1,0.3,0.0021)  

(0.05,0.35,0.0000) (0.1,0.35,0.0000) (0.15,0.35,0.0001) (0.2,0.35,0.0010) (0.25,0.35,0.0083) (0.3,0.35,0.0540) (0.35,0.35,0.2673) (0.4,0.35,1.0019) (0.45,0.35,2.8473) (0.5,0.35,6.1338) (0.55,0.35,10.0168) (0.6,0.35,12.4001) (0.65,0.35,11.6363) (0.7,0.35,8.2776) (0.75,0.35,4.4636) (0.8,0.35,1.8246) (0.85,0.35,0.5654) (0.9,0.35,0.1328) (0.95,0.35,0.0236) (1,0.35,0.0032)  

(0.05,0.4,0.0000) (0.1,0.4,0.0000) (0.15,0.4,0.0001) (0.2,0.4,0.0011) (0.25,0.4,0.0097) (0.3,0.4,0.0634) (0.35,0.4,0.3137) (0.4,0.4,1.1759) (0.45,0.4,3.3418) (0.5,0.4,7.1992) (0.55,0.4,11.7567) (0.6,0.4,14.5539) (0.65,0.4,13.6574) (0.7,0.4,9.7153) (0.75,0.4,5.2389) (0.8,0.4,2.1415) (0.85,0.4,0.6636) (0.9,0.4,0.1559) (0.95,0.4,0.0278) (1,0.4,0.0037)  

(0.05,0.45,0.0000) (0.1,0.45,0.0000) (0.15,0.45,0.0001) (0.2,0.45,0.0010) (0.25,0.45,0.0086) (0.3,0.45,0.0564) (0.35,0.45,0.2791) (0.4,0.45,1.0462) (0.45,0.45,2.9733) (0.5,0.45,6.4053) (0.55,0.45,10.4601) (0.6,0.45,12.9488) (0.65,0.45,12.1512) (0.7,0.45,8.6439) (0.75,0.45,4.6611) (0.8,0.45,1.9053) (0.85,0.45,0.5904) (0.9,0.45,0.1387) (0.95,0.45,0.0247) (1,0.45,0.0033)  

(0.05,0.5,0.0000) (0.1,0.5,0.0000) (0.15,0.5,0.0001) (0.2,0.5,0.0007) (0.25,0.5,0.0058) (0.3,0.5,0.0381) (0.35,0.5,0.1882) (0.4,0.5,0.7056) (0.45,0.5,2.0053) (0.5,0.5,4.3200) (0.55,0.5,7.0547) (0.6,0.5,8.7332) (0.65,0.5,8.1953) (0.7,0.5,5.8298) (0.75,0.5,3.1437) (0.8,0.5,1.2851) (0.85,0.5,0.3982) (0.9,0.5,0.0935) (0.95,0.5,0.0167) (1,0.5,0.0022)  

(0.05,0.55,0.0000) (0.1,0.55,0.0000) (0.15,0.55,0.0000) (0.2,0.55,0.0003) (0.25,0.55,0.0030) (0.3,0.55,0.0195) (0.35,0.55,0.0962) (0.4,0.55,0.3608) (0.45,0.55,1.0252) (0.5,0.55,2.2087) (0.55,0.55,3.6068) (0.6,0.55,4.4650) (0.65,0.55,4.1900) (0.7,0.55,2.9806) (0.75,0.55,1.6072) (0.8,0.55,0.6570) (0.85,0.55,0.2036) (0.9,0.55,0.0478) (0.95,0.55,0.0085) (1,0.55,0.0011)  

(0.05,0.6,0.0000) (0.1,0.6,0.0000) (0.15,0.6,0.0000) (0.2,0.6,0.0001) (0.25,0.6,0.0012) (0.3,0.6,0.0075) (0.35,0.6,0.0373) (0.4,0.6,0.1398) (0.45,0.6,0.3973) (0.5,0.6,0.8560) (0.55,0.6,1.3979) (0.6,0.6,1.7305) (0.65,0.6,1.6239) (0.7,0.6,1.1552) (0.75,0.6,0.6229) (0.8,0.6,0.2546) (0.85,0.6,0.0789) (0.9,0.6,0.0185) (0.95,0.6,0.0033) (1,0.6,0.0004)  

(0.05,0.65,0.0000) (0.1,0.65,0.0000) (0.15,0.65,0.0000) (0.2,0.65,0.0000) (0.25,0.65,0.0003) (0.3,0.65,0.0022) (0.35,0.65,0.0110) (0.4,0.65,0.0411) (0.45,0.65,0.1167) (0.5,0.65,0.2515) (0.55,0.65,0.4107) (0.6,0.65,0.5084) (0.65,0.65,0.4771) (0.7,0.65,0.3394) (0.75,0.65,0.1830) (0.8,0.65,0.0748) (0.85,0.65,0.0232) (0.9,0.65,0.0054) (0.95,0.65,0.0010) (1,0.65,0.0001)  

(0.05,0.7,0.0000) (0.1,0.7,0.0000) (0.15,0.7,0.0000) (0.2,0.7,0.0000) (0.25,0.7,0.0001) (0.3,0.7,0.0005) (0.35,0.7,0.0024) (0.4,0.7,0.0091) (0.45,0.7,0.0260) (0.5,0.7,0.0560) (0.55,0.7,0.0915) (0.6,0.7,0.1132) (0.65,0.7,0.1062) (0.7,0.7,0.0756) (0.75,0.7,0.0408) (0.8,0.7,0.0167) (0.85,0.7,0.0052) (0.9,0.7,0.0012) (0.95,0.7,0.0002) (1,0.7,0.0000)  

(0.05,0.75,0.0000) (0.1,0.75,0.0000) (0.15,0.75,0.0000) (0.2,0.75,0.0000) (0.25,0.75,0.0000) (0.3,0.75,0.0001) (0.35,0.75,0.0004) (0.4,0.75,0.0015) (0.45,0.75,0.0044) (0.5,0.75,0.0095) (0.55,0.75,0.0154) (0.6,0.75,0.0191) (0.65,0.75,0.0179) (0.7,0.75,0.0128) (0.75,0.75,0.0069) (0.8,0.75,0.0028) (0.85,0.75,0.0009) (0.9,0.75,0.0002) (0.95,0.75,0.0000) (1,0.75,0.0000)  

(0.05,0.8,0.0000) (0.1,0.8,0.0000) (0.15,0.8,0.0000) (0.2,0.8,0.0000) (0.25,0.8,0.0000) (0.3,0.8,0.0000) (0.35,0.8,0.0001) (0.4,0.8,0.0002) (0.45,0.8,0.0006) (0.5,0.8,0.0012) (0.55,0.8,0.0020) (0.6,0.8,0.0024) (0.65,0.8,0.0023) (0.7,0.8,0.0016) (0.75,0.8,0.0009) (0.8,0.8,0.0004) (0.85,0.8,0.0001) (0.9,0.8,0.0000) (0.95,0.8,0.0000) (1,0.8,0.0000)  

(0.05,0.85,0.0000) (0.1,0.85,0.0000) (0.15,0.85,0.0000) (0.2,0.85,0.0000) (0.25,0.85,0.0000) (0.3,0.85,0.0000) (0.35,0.85,0.0000) (0.4,0.85,0.0000) (0.45,0.85,0.0001) (0.5,0.85,0.0001) (0.55,0.85,0.0002) (0.6,0.85,0.0002) (0.65,0.85,0.0002) (0.7,0.85,0.0002) (0.75,0.85,0.0001) (0.8,0.85,0.0000) (0.85,0.85,0.0000) (0.9,0.85,0.0000) (0.95,0.85,0.0000) (1,0.85,0.0000)  

(0.05,0.9,0.0000) (0.1,0.9,0.0000) (0.15,0.9,0.0000) (0.2,0.9,0.0000) (0.25,0.9,0.0000) (0.3,0.9,0.0000) (0.35,0.9,0.0000) (0.4,0.9,0.0000) (0.45,0.9,0.0000) (0.5,0.9,0.0000) (0.55,0.9,0.0000) (0.6,0.9,0.0000) (0.65,0.9,0.0000) (0.7,0.9,0.0000) (0.75,0.9,0.0000) (0.8,0.9,0.0000) (0.85,0.9,0.0000) (0.9,0.9,0.0000) (0.95,0.9,0.0000) (1,0.9,0.0000)  

(0.05,0.95,0.0000) (0.1,0.95,0.0000) (0.15,0.95,0.0000) (0.2,0.95,0.0000) (0.25,0.95,0.0000) (0.3,0.95,0.0000) (0.35,0.95,0.0000) (0.4,0.95,0.0000) (0.45,0.95,0.0000) (0.5,0.95,0.0000) (0.55,0.95,0.0000) (0.6,0.95,0.0000) (0.65,0.95,0.0000) (0.7,0.95,0.0000) (0.75,0.95,0.0000) (0.8,0.95,0.0000) (0.85,0.95,0.0000) (0.9,0.95,0.0000) (0.95,0.95,0.0000) (1,0.95,0.0000) 

(0.05,1,0.0000) (0.1,1,0.0000) (0.15,1,0.0000) (0.2,1,0.0000) (0.25,1,0.0000) (0.3,1,0.0000) (0.35,1,0.0000) (0.4,1,0.0000) (0.45,1,0.0000) (0.5,1,0.0000) (0.55,1,0.0000) (0.6,1,0.0000) (0.65,1,0.0000) (0.7,1,0.0000) (0.75,1,0.0000) (0.8,1,0.0000) (0.85,1,0.0000) (0.9,1,0.0000) (0.95,1,0.0000) (1,1,0.0000)  
};
\end{axis}
\end{tikzpicture}
\caption{
Example signal from class $\mathbb{P}$.
\vspace{5pt}
}
\label{subfig:App:Syn2:Res:ExP}
\end{subfigure}
\hfill
\begin{subfigure}[t]{0.45\columnwidth}
\centering
\scriptsize
\begin{tikzpicture} 
\begin{axis}[
width=\figurewidth,
height=\figureheight,
scale only axis,
xmin=-0,
xmax=1,
ymin=0,
ymax=1,
zmin=-10,
zmax=12,
axis background/.style={fill=white!100},
]
\addplot3[surf] coordinates{
(0.05,0.05,-0.0000) (0.1,0.05,-0.0000) (0.15,0.05,-0.0000) (0.2,0.05,-0.0000) (0.25,0.05,-0.0000) (0.3,0.05,-0.0000) (0.35,0.05,-0.0000) (0.4,0.05,0.0000) (0.45,0.05,0.0000) (0.5,0.05,0.0000) (0.55,0.05,0.0003) (0.6,0.05,0.0021) (0.65,0.05,0.0097) (0.7,0.05,0.0342) (0.75,0.05,0.0915) (0.8,0.05,0.1860) (0.85,0.05,0.2865) (0.9,0.05,0.3345) (0.95,0.05,0.2961) (1,0.05,0.1987)  

(0.05,0.1,-0.0000) (0.1,0.1,-0.0000) (0.15,0.1,-0.0000) (0.2,0.1,-0.0000) (0.25,0.1,-0.0000) (0.3,0.1,-0.0000) (0.35,0.1,-0.0000) (0.4,0.1,-0.0000) (0.45,0.1,0.0000) (0.5,0.1,0.0001) (0.55,0.1,0.0011) (0.6,0.1,0.0070) (0.65,0.1,0.0328) (0.7,0.1,0.1160) (0.75,0.1,0.3108) (0.8,0.1,0.6316) (0.85,0.1,0.9729) (0.9,0.1,1.1360) (0.95,0.1,1.0055) (1,0.1,0.6746)  

(0.05,0.15,-0.0000) (0.1,0.15,-0.0000) (0.15,0.15,-0.0000) (0.2,0.15,-0.0000) (0.25,0.15,-0.0000) (0.3,0.15,-0.0000) (0.35,0.15,-0.0000) (0.4,0.15,-0.0000) (0.45,0.15,-0.0000) (0.5,0.15,0.0003) (0.55,0.15,0.0029) (0.6,0.15,0.0180) (0.65,0.15,0.0844) (0.7,0.15,0.2985) (0.75,0.15,0.8001) (0.8,0.15,1.6258) (0.85,0.15,2.5042) (0.9,0.15,2.9239) (0.95,0.15,2.5880) (1,0.15,1.7365)  

(0.05,0.2,-0.0000) (0.1,0.2,-0.0000) (0.15,0.2,-0.0000) (0.2,0.2,-0.0000) (0.25,0.2,-0.0000) (0.3,0.2,-0.0000) (0.35,0.2,-0.0001) (0.4,0.2,-0.0003) (0.45,0.2,-0.0004) (0.5,0.2,0.0000) (0.55,0.2,0.0050) (0.6,0.2,0.0348) (0.65,0.2,0.1644) (0.7,0.2,0.5823) (0.75,0.2,1.5611) (0.8,0.2,3.1722) (0.85,0.2,4.8861) (0.9,0.2,5.7052) (0.95,0.2,5.0497) (1,0.2,3.3882)  

(0.05,0.25,-0.0000) (0.1,0.25,-0.0000) (0.15,0.25,-0.0000) (0.2,0.25,-0.0000) (0.25,0.25,-0.0001) (0.3,0.25,-0.0003) (0.35,0.25,-0.0011) (0.4,0.25,-0.0025) (0.45,0.25,-0.0043) (0.5,0.25,-0.0049) (0.55,0.25,0.0024) (0.6,0.25,0.0475) (0.65,0.25,0.2409) (0.7,0.25,0.8602) (0.75,0.25,2.3087) (0.8,0.25,4.6918) (0.85,0.25,7.2270) (0.9,0.25,8.4385) (0.95,0.25,7.4690) (1,0.25,5.0114)  

(0.05,0.3,-0.0000) (0.1,0.3,-0.0000) (0.15,0.3,-0.0000) (0.2,0.3,-0.0001) (0.25,0.3,-0.0006) (0.3,0.3,-0.0022) (0.35,0.3,-0.0069) (0.4,0.3,-0.0162) (0.45,0.3,-0.0287) (0.5,0.3,-0.0377) (0.55,0.3,-0.0301) (0.6,0.3,0.0279) (0.65,0.3,0.2552) (0.7,0.3,0.9579) (0.75,0.3,2.5863) (0.8,0.3,5.2600) (0.85,0.3,8.1030) (0.9,0.3,9.4614) (0.95,0.3,8.3744) (1,0.3,5.6189)  

(0.05,0.35,-0.0000) (0.1,0.35,-0.0000) (0.15,0.35,-0.0001) (0.2,0.35,-0.0005) (0.25,0.35,-0.0027) (0.3,0.35,-0.0111) (0.35,0.35,-0.0343) (0.4,0.35,-0.0804) (0.45,0.35,-0.1427) (0.5,0.35,-0.1912) (0.55,0.35,-0.1881) (0.6,0.35,-0.1020) (0.65,0.35,0.1432) (0.7,0.35,0.7814) (0.75,0.35,2.1872) (0.8,0.35,4.4679) (0.85,0.35,6.8865) (0.9,0.35,8.0416) (0.95,0.35,7.1178) (1,0.35,4.7758)  

(0.05,0.4,-0.0000) (0.1,0.4,-0.0000) (0.15,0.4,-0.0003) (0.2,0.4,-0.0019) (0.25,0.4,-0.0102) (0.3,0.4,-0.0417) (0.35,0.4,-0.1288) (0.4,0.4,-0.3018) (0.45,0.4,-0.5358) (0.5,0.4,-0.7207) (0.55,0.4,-0.7309) (0.6,0.4,-0.5373) (0.65,0.4,-0.1843) (0.7,0.4,0.3805) (0.75,0.4,1.3678) (0.8,0.4,2.8681) (0.85,0.4,4.4349) (0.9,0.4,5.1809) (0.95,0.4,4.5859) (1,0.4,3.0770)  

(0.05,0.45,-0.0000) (0.1,0.45,-0.0001) (0.15,0.45,-0.0008) (0.2,0.45,-0.0054) (0.25,0.45,-0.0291) (0.3,0.45,-0.1186) (0.35,0.45,-0.3664) (0.4,0.45,-0.8584) (0.45,0.45,-1.5243) (0.5,0.45,-2.0516) (0.55,0.45,-2.0912) (0.6,0.45,-1.6039) (0.65,0.45,-0.8766) (0.7,0.45,-0.1638) (0.75,0.45,0.5502) (0.8,0.45,1.3707) (0.85,0.45,2.1602) (0.9,0.45,2.5295) (0.95,0.45,2.2397) (1,0.45,1.5028)  

(0.05,0.5,-0.0000) (0.1,0.5,-0.0002) (0.15,0.5,-0.0016) (0.2,0.5,-0.0117) (0.25,0.5,-0.0627) (0.3,0.5,-0.2557) (0.35,0.5,-0.7902) (0.4,0.5,-1.8510) (0.45,0.5,-3.2869) (0.5,0.5,-4.4244) (0.55,0.5,-4.5139) (0.6,0.5,-3.4865) (0.65,0.5,-2.0207) (0.7,0.5,-0.8146) (0.75,0.5,-0.0503) (0.8,0.5,0.4426) (0.85,0.5,0.7872) (0.9,0.5,0.9347) (0.95,0.5,0.8290) (1,0.5,0.5564)  

(0.05,0.55,-0.0000) (0.1,0.55,-0.0003) (0.15,0.55,-0.0027) (0.2,0.55,-0.0191) (0.25,0.55,-0.1025) (0.3,0.55,-0.4179) (0.35,0.55,-1.2916) (0.4,0.55,-3.0255) (0.45,0.55,-5.3726) (0.5,0.55,-7.2321) (0.55,0.55,-7.3795) (0.6,0.55,-5.7068) (0.65,0.55,-3.3397) (0.7,0.55,-1.4610) (0.75,0.55,-0.4294) (0.8,0.55,0.0182) (0.85,0.55,0.2004) (0.9,0.55,0.2593) (0.95,0.55,0.2323) (1,0.55,0.1561)  

(0.05,0.6,-0.0000) (0.1,0.6,-0.0004) (0.15,0.6,-0.0033) (0.2,0.6,-0.0236) (0.25,0.6,-0.1270) (0.3,0.6,-0.5179) (0.35,0.6,-1.6003) (0.4,0.6,-3.7489) (0.45,0.6,-6.6571) (0.5,0.6,-8.9612) (0.55,0.6,-9.1441) (0.6,0.6,-7.0729) (0.65,0.6,-4.1459) (0.7,0.6,-1.8378) (0.75,0.6,-0.6059) (0.8,0.6,-0.1276) (0.85,0.6,0.0172) (0.9,0.6,0.0514) (0.95,0.6,0.0490) (1,0.6,0.0332)  

(0.05,0.65,-0.0000) (0.1,0.65,-0.0003) (0.15,0.65,-0.0031) (0.2,0.65,-0.0222) (0.25,0.65,-0.1193) (0.3,0.65,-0.4864) (0.35,0.65,-1.5032) (0.4,0.65,-3.5212) (0.45,0.65,-6.2529) (0.5,0.65,-8.4171) (0.55,0.65,-8.5889) (0.6,0.65,-6.6437) (0.65,0.65,-3.8954) (0.7,0.65,-1.7307) (0.75,0.65,-0.5810) (0.8,0.65,-0.1440) (0.85,0.65,-0.0211) (0.9,0.65,0.0048) (0.95,0.65,0.0075) (1,0.65,0.0053)  

(0.05,0.7,-0.0000) (0.1,0.7,-0.0002) (0.15,0.7,-0.0022) (0.2,0.7,-0.0158) (0.25,0.7,-0.0850) (0.3,0.7,-0.3463) (0.35,0.7,-1.0703) (0.4,0.7,-2.5072) (0.45,0.7,-4.4522) (0.5,0.7,-5.9931) (0.55,0.7,-6.1155) (0.6,0.7,-4.7305) (0.65,0.7,-2.7738) (0.7,0.7,-1.2328) (0.75,0.7,-0.4151) (0.8,0.7,-0.1055) (0.85,0.7,-0.0196) (0.9,0.7,-0.0019) (0.95,0.7,0.0006) (1,0.7,0.0006)  

(0.05,0.75,-0.0000) (0.1,0.75,-0.0001) (0.15,0.75,-0.0012) (0.2,0.75,-0.0085) (0.25,0.75,-0.0459) (0.3,0.75,-0.1869) (0.35,0.75,-0.5777) (0.4,0.75,-1.3533) (0.45,0.75,-2.4031) (0.5,0.75,-3.2348) (0.55,0.75,-3.3008) (0.6,0.75,-2.5533) (0.65,0.75,-1.4972) (0.7,0.75,-0.6655) (0.75,0.75,-0.2242) (0.8,0.75,-0.0572) (0.85,0.75,-0.0110) (0.9,0.75,-0.0015) (0.95,0.75,-0.0001) (1,0.75,0.0000) 
 
(0.05,0.8,-0.0000) (0.1,0.8,-0.0001) (0.15,0.8,-0.0005) (0.2,0.8,-0.0035) (0.25,0.8,-0.0188) (0.3,0.8,-0.0765) (0.35,0.8,-0.2364) (0.4,0.8,-0.5537) (0.45,0.8,-0.9832) (0.5,0.8,-1.3235) (0.55,0.8,-1.3505) (0.6,0.8,-1.0447) (0.65,0.8,-0.6126) (0.7,0.8,-0.2723) (0.75,0.8,-0.0917) (0.8,0.8,-0.0234) (0.85,0.8,-0.0045) (0.9,0.8,-0.0007) (0.95,0.8,-0.0001) (1,0.8,-0.0000)  

(0.05,0.85,-0.0000) (0.1,0.85,-0.0000) (0.15,0.85,-0.0002) (0.2,0.85,-0.0011) (0.25,0.85,-0.0058) (0.3,0.85,-0.0237) (0.35,0.85,-0.0733) (0.4,0.85,-0.1717) (0.45,0.85,-0.3050) (0.5,0.85,-0.4105) (0.55,0.85,-0.4189) (0.6,0.85,-0.3240) (0.65,0.85,-0.1900) (0.7,0.85,-0.0845) (0.75,0.85,-0.0285) (0.8,0.85,-0.0073) (0.85,0.85,-0.0014) (0.9,0.85,-0.0002) (0.95,0.85,-0.0000) (1,0.85,-0.0000)  

(0.05,0.9,-0.0000) (0.1,0.9,-0.0000) (0.15,0.9,-0.0000) (0.2,0.9,-0.0003) (0.25,0.9,-0.0014) (0.3,0.9,-0.0056) (0.35,0.9,-0.0172) (0.4,0.9,-0.0404) (0.45,0.9,-0.0717) (0.5,0.9,-0.0965) (0.55,0.9,-0.0985) (0.6,0.9,-0.0762) (0.65,0.9,-0.0447) (0.7,0.9,-0.0199) (0.75,0.9,-0.0067) (0.8,0.9,-0.0017) (0.85,0.9,-0.0003) (0.9,0.9,-0.0000) (0.95,0.9,-0.0000) (1,0.9,-0.0000)  

(0.05,0.95,-0.0000) (0.1,0.95,-0.0000) (0.15,0.95,-0.0000) (0.2,0.95,-0.0000) (0.25,0.95,-0.0002) (0.3,0.95,-0.0010) (0.35,0.95,-0.0031) (0.4,0.95,-0.0072) (0.45,0.95,-0.0128) (0.5,0.95,-0.0172) (0.55,0.95,-0.0176) (0.6,0.95,-0.0136) (0.65,0.95,-0.0080) (0.7,0.95,-0.0035) (0.75,0.95,-0.0012) (0.8,0.95,-0.0003) (0.85,0.95,-0.0001) (0.9,0.95,-0.0000) (0.95,0.95,-0.0000) (1,0.95,-0.0000) 
 
(0.05,1,-0.0000) (0.1,1,-0.0000) (0.15,1,-0.0000) (0.2,1,-0.0000) (0.25,1,-0.0000) (0.3,1,-0.0001) (0.35,1,-0.0004) (0.4,1,-0.0010) (0.45,1,-0.0017) (0.5,1,-0.0023) (0.55,1,-0.0024) (0.6,1,-0.0018) (0.65,1,-0.0011) (0.7,1,-0.0005) (0.75,1,-0.0002) (0.8,1,-0.0000) (0.85,1,-0.0000) (0.9,1,-0.0000) (0.95,1,-0.0000) (1,1,-0.0000)  
};
\end{axis}
\end{tikzpicture}
\caption{
Example signal from class $\mathbb{Q}$.
\vspace{5pt}
}
\label{subfig:App:Syn2:Res:ExQ}
\end{subfigure}

\begin{subfigure}[t]{0.45\columnwidth}
\centering
\scriptsize
%
%
%
%
\begin{tikzpicture}

\begin{axis}[%
width=\figurewidth,
height=\figureheight,
scale only axis,
xmin=-60,
xmax=80,
ymin=-50,
ymax=55,
xticklabels={},
yticklabels={},
axis background/.style={fill=white!100},
]
\addplot [
color=black,
mark size=3.0pt,
only marks,
mark=x,
mark options={solid},
forget plot
]
table[row sep=crcr]{
-6.31135785498803 -6.14244262484336\\
24.6910167867093 20.6104562448634\\
-1.93657834631702 -6.20333857836433\\
12.1495557894501 -3.91556288809709\\
5.7123569875044 5.49136711510257\\
-7.80206521799292 14.3344676768935\\
-6.00491112830441 -6.66391627716345\\
5.22830767447651 -3.33192925500342\\
-1.24507345624096 10.8685457780819\\
5.67298003331746 -7.31525987032597\\
-22.2454495985291 -13.4766720715846\\
15.2192066456439 -2.10349046997423\\
22.456548512998 -21.8530250452916\\
0.284037474504507 -6.59060190026304\\
10.2681863638563 -1.49733273291938\\
5.10267412981509 13.0809147839579\\
-8.4206293238946 -2.13869834983932\\
-1.77357127664357 -0.0416515581396551\\
7.20090405492137 -10.7106960539624\\
-1.42002880070554 18.8292439057644\\
-19.9744795303259 -6.41422938216705\\
-5.64559270339599 -2.85184528520309\\
-4.48287739028226 13.4676976668024\\
13.0249119624032 6.1428743840772\\
9.94126988463689 -17.6239540951628\\
};
\addplot [
color=red,
mark size=3.0pt,
only marks,
mark=o,
mark options={solid},
forget plot
]
table[row sep=crcr]{
-6.77889777366056 -16.0371808152029\\
0.149405345610203 -0.729749246771786\\
10.6947935728085 20.5277473966159\\
-10.6690260235816 -29.1500095032172\\
-6.30567210286216 2.02822028626894\\
-25.8958550392436 -47.8522564432107\\
40.6828226691373 -20.0600723642786\\
16.5950209785373 48.9386678118353\\
-55.5331400004644 -18.3126704808602\\
-51.5848113990905 8.85479405942175\\
-21.7239838723814 4.81956923784302\\
-9.86503138849104 -19.3802072855216\\
14.9304709797396 -45.1028139364833\\
-5.99942517271993 49.627739701935\\
63.005530393976 -4.73571992616368\\
3.55559781761298 -19.4528063514406\\
18.4319889445409 2.3201172788039\\
-10.6569874581834 30.2543252973911\\
3.37491995328881 -38.3758341835103\\
-47.9874175284495 30.0889900869689\\
29.2289039538283 30.7709406050794\\
0.161610107205831 0.777792444355731\\
-15.6467145335129 5.96250870450955\\
44.9325696282556 11.534805141967\\
-26.7860137245168 28.7321813664268\\
};
\end{axis}
\end{tikzpicture}%
\caption{ 
MDS projection in $L^2$.
\vspace{5pt}
}
\label{subfig:App:Syn2:Res:ResL2}
\end{subfigure}
\hfill
\begin{subfigure}[t]{0.45\columnwidth}
\centering
\scriptsize
%
%
%
%
\begin{tikzpicture}

\begin{axis}[%
width=\figurewidth,
height=\figureheight,
scale only axis,
xmin=-4,
xmax=4,
ymin=-3,
ymax=3,
xticklabels={},
yticklabels={},
axis background/.style={fill=white!100},
]
\addplot [
color=black,
mark size=3.0pt,
only marks,
mark=x,
mark options={solid},
forget plot
]
table[row sep=crcr]{
0.564027604102893 -0.417184202377032\\
-0.321118794594007 1.28688035136788\\
-1.19510629122325 -0.502225000269003\\
2.09075541867236 0.168455412941048\\
-2.12164955722755 0.450174786566569\\
0.443750745778548 1.54543709605433\\
0.302868890760176 -2.09589598995644\\
-1.38777559115318 0.216519835094942\\
-0.659854141100247 0.842770052568285\\
0.676200920622512 0.533339698401985\\
-0.00464600405928974 -2.04086832624587\\
-0.340569898520157 -0.578924855116549\\
-0.504384240433287 -0.31733170554761\\
0.469217526975547 2.67283869019862\\
-0.589941474886471 1.08448376085619\\
-0.0291854735272842 -1.35819138911224\\
-0.504846838308412 1.23142779932705\\
-0.385061341214512 -1.09960416571421\\
0.932646554818231 1.38109751748097\\
0.178234266732425 -0.880783232640879\\
0.870358904951322 0.657030730905122\\
-1.99324511884437 -0.812591319897227\\
0.205392063586043 0.827493164466962\\
2.73596497934668 0.0193889515863756\\
-0.615163956757587 0.532162010230745\\
};
\addplot [
color=red,
mark size=3.0pt,
only marks,
mark=o,
mark options={solid},
forget plot
]
table[row sep=crcr]{
3.40838900658273 -1.98894332779523\\
-0.134860250468551 -1.51020453727231\\
1.69955146387808 0.845955490692347\\
0.18242627953059 -0.0206940157949349\\
-2.01725485562635 1.35213052750111\\
-1.7082886323803 -0.171129162024683\\
1.51268581277847 1.30819317715218\\
-0.732678039288935 -1.6359006383234\\
-1.17720234280117 -0.636881031965592\\
-1.07917377602645 -0.838889120881124\\
-0.579276437279044 -0.216204542037651\\
-3.26483244758034 0.5284339607282\\
1.53999672055736 -0.0765791421233327\\
-1.57347067215728 0.747926883390671\\
-1.18356989690519 0.673663180492062\\
1.46166229558386 0.361950566176032\\
-0.803318274752091 -0.961796701970403\\
0.0546578182076235 0.176407270723598\\
1.48007769988775 -2.26047195909382\\
3.72111118466578 1.74721854773608\\
-1.1191580138333 1.84078584028733\\
-0.280366053979734 0.161723644707678\\
-2.27720453505543 -1.56995085756273\\
2.05279975823495 -0.750428454443651\\
2.00042703372986 -0.452215269468466\\
};
\end{axis}
\end{tikzpicture}%
\caption{
MDS projection in OT.
\vspace{5pt}
}
\label{subfig:App:Syn2:Res:ResOT}
\end{subfigure}

\begin{subfigure}[b]{0.45\columnwidth}
\centering
\scriptsize
%
%
%
%
\begin{tikzpicture}

\begin{axis}[%
width=\figurewidth,
height=\figureheight,
scale only axis,
xmin=-1,
xmax=2,
ymin=-1,
ymax=2,
xticklabels={},
yticklabels={},
axis background/.style={fill=white!100},
]
\addplot [
color=black,
mark size=3.0pt,
only marks,
mark=x,
mark options={solid},
forget plot
]
table[row sep=crcr]{
-0.856724109849194 0.426599039557051\\
-0.840426956525227 -0.0127673420783643\\
-0.854563981330797 -0.521473415321458\\
-0.941250629200751 -0.17971128896854\\
-0.84519265978787 -0.250243929009363\\
-0.846687888287147 0.0968869346227454\\
-0.874967463782983 -0.499301340336262\\
-0.839502393455708 -0.0320605477428219\\
-0.854938853694703 -0.336935698490972\\
-0.865907802523163 -0.613291729574617\\
-0.878773121581934 -0.72837031944141\\
-0.853128094662456 -0.528786830630475\\
-0.849025089645454 0.421309090148979\\
-0.91111348232125 -0.40984397709029\\
-0.895931417188105 0.122234620522034\\
-0.84644039246024 0.00358477308651439\\
-0.840475726859809 0.0320524002284745\\
-0.842569928018299 -0.33732296648721\\
-0.858104089492929 -0.433039655539109\\
-0.853727716616144 0.351940038187166\\
-0.849805756659104 0.22141120814018\\
-0.864491143495144 -0.751804034924762\\
-0.869299498390074 0.838267540847359\\
-0.864101647323272 -0.575227449213056\\
-0.850043780202155 0.288515229638664\\
};
\addplot [
color=red,
mark size=3.0pt,
only marks,
mark=o,
mark options={solid},
forget plot
]
table[row sep=crcr]{
1.66491867560407 -0.471626351884792\\
0.697667110416723 0.26457709624325\\
0.695703872934792 0.220533722827762\\
-0.226345542440816 1.26251134731013\\
0.687617262919893 0.357768795026737\\
1.10523547630331 -0.207202737260356\\
0.641766842261876 0.236449128198981\\
0.500912770394017 0.450352684276776\\
0.341683930244662 0.385008578847198\\
0.546890968202835 0.352232059399351\\
0.794747231720427 0.110101644550514\\
1.40690692665711 -0.516473114983877\\
0.622273775890946 0.38243964169272\\
1.05975311091349 -0.109270832881773\\
1.2160793591237 -0.221783089554769\\
0.980273083884783 0.164958875701287\\
1.12176917288322 -0.138017435111002\\
-0.56362621123104 1.50891666721767\\
1.09763395470177 -0.0119010668308892\\
1.81137262165896 -0.240026758088133\\
1.02969417260384 -0.0810691276433668\\
0.578721583698698 0.407836329355028\\
0.921496273399661 -0.235620303365298\\
1.67081894406494 -0.555866312681881\\
1.14322825654203 0.0925502095082761\\
};
\end{axis}
\end{tikzpicture}%
\caption{
MDS projection in $TL^2$.
\vspace{20pt}
}
\label{subfig:App:Syn2:Res:ResTL2}
\end{subfigure}
\hfill
\begin{subfigure}[b]{0.45\columnwidth}
\centering
\scriptsize
%
%
%
%
\begin{tikzpicture}

\begin{axis}[%
width=\figurewidth,
height=\figureheight,
scale only axis,
xmin=-2.7,
xmax=27,
ymin=-0.035,
ymax=0.27,
hide axis,
axis background/.style={fill=white!100},
]
\draw [->] (axis cs: 0,0) -- (axis cs: 26,0) node[below] {$k$};
\draw [->] (axis cs: 0,0) -- (axis cs: 0,0.25) node[left] {$S_{\rho}$};
\draw[-] (axis cs: 0,0) -- ++(2pt,0) node[left,xshift=-2pt] {0};
\draw[-] (axis cs: 0,0.1) -- ++(2pt,0) node[left,xshift=-2pt] {0.1};
\draw[-] (axis cs: 0,0.2) -- ++(2pt,0) node[left,xshift=-2pt] {0.2};
\draw[-] (axis cs: 0,0) -- ++(0,2pt) node[below,yshift=-2pt] {0};
\draw[-] (axis cs: 5,0) -- ++(0,2pt) node[below,yshift=-2pt] {5};
\draw[-] (axis cs: 10,0) -- ++(0,2pt) node[below,yshift=-2pt] {10};
\draw[-] (axis cs: 15,0) -- ++(0,2pt) node[below,yshift=-2pt] {15};
\draw[-] (axis cs: 20,0) -- ++(0,2pt) node[below,yshift=-2pt] {20};
\draw[-] (axis cs: 25,0) -- ++(0,2pt);
\addplot [
color=blue,
solid,
line width = 1.2pt,
forget plot
]
table[row sep=crcr]{
2 0.0346964662912208\\
3 0.0203485455521482\\
4 0.0125866203796491\\
5 0.00738709378047515\\
6 0.00359734536457319\\
7 0.00241526677217222\\
8 0.00150304198073673\\
9 0.00114389645896386\\
10 0.000964260351496819\\
11 0.00198035900661846\\
12 0.00294282313096406\\
13 0.00256810346866047\\
14 0.00204394879241351\\
15 0.00234073828822097\\
16 0.00207526800192084\\
17 0.00244404235012505\\
18 0.00237733783193999\\
19 0.00209858719730864\\
20 0.00215500840378705\\
21 0.00199180746257504\\
22 0.00205032116095583\\
23 0.00180334118566892\\
24 0.00192281281360931\\
25 0.00195746607153359\\
};
\addplot [
color=red,
solid,
line width = 1.2pt,
forget plot
]
table[row sep=crcr]{
2 0.245492896981252\\
3 0.160550014503084\\
4 0.115978859960276\\
5 0.0920203125666073\\
6 0.0742573301423668\\
7 0.060071874255279\\
8 0.0495476963185686\\
9 0.0406739903595095\\
10 0.033832638813156\\
11 0.0275376486960565\\
12 0.0224070382006491\\
13 0.0173831505083549\\
14 0.0147012025984232\\
15 0.0123348782160155\\
16 0.00990403790714443\\
17 0.00844695570901495\\
18 0.00682684194132428\\
19 0.00553987667389768\\
20 0.00468201367332377\\
21 0.00357233009488795\\
22 0.00244597699481588\\
23 0.00184495586810335\\
24 0.00138117084099432\\
25 0.00146453624636698\\
};
\addplot [
color=black,
solid,
line width = 1.2pt,
forget plot
]
table[row sep=crcr]{
2 0.133656835958723\\
3 0.0880386082594357\\
4 0.0595741801050114\\
5 0.0433688948216874\\
6 0.0275124863303503\\
7 0.0229131065624372\\
8 0.0188399125862293\\
9 0.01566260594126\\
10 0.0136313536139367\\
11 0.0121168726894572\\
12 0.010766804800258\\
13 0.00953418111182527\\
14 0.00846898565508316\\
15 0.00739435666453438\\
16 0.00645320354305851\\
17 0.00562780638299214\\
18 0.0050384644730781\\
19 0.00455751380798328\\
20 0.0040405979619239\\
21 0.00365483294276137\\
22 0.00335118849604905\\
23 0.00275541531946548\\
24 0.00237608701727161\\
25 0.0020631878742747\\
};
\end{axis}
\end{tikzpicture}%
\vspace{5pt}
\caption{
Dependence of the stress $S_{\rho}$ on the dimension of the projected space $k$ for $L^2$ (red), OT (black) and $TL^2$ (blue).
}
\label{subfig:App:Syn2:Res:StressRes}
\end{subfigure}
\caption{
Example signals and results for the synthetic data in Section~\ref{subsec:App:Syn2}.
}
\label{fig:App:Syn2:Res}
\end{figbox}
\end{figure*}

\paragraph*{Data Sets.}
The data set consists of two dimensional images simulated from the following classes
\begin{align*}
\mathbb{P} & = \Bigg\{ p\lfloor_{[0,1]^2} \, : \, p(x) = \alpha \phi(x| \gamma, \sigma), \, \gamma \sim \mathrm{unif}([0,1]^2), \\
 & \quad \quad \alpha \sim \mathrm{unif}([0.5,1]) \Bigg\}
\end{align*}
\begin{align*}
\mathbb{Q} & = \Bigg\{ q\lfloor_{[0,1]^2} \, : \, q(x) = \alpha \phi(x| \gamma_1, \sigma) - \alpha \phi(x| \gamma_2, \sigma), \\
 & \quad \quad \gamma_1,\gamma_2 \iid \mathrm{unif}([0,1]^2), \, \alpha \sim \mathrm{unif}([0.5,1]) \Bigg\}
\end{align*}
where $\phi(\cdot|\gamma,\sigma)$ is the multivariate normal pdf with mean $\gamma\in \mathbb{R}^2$ and co-variance $\sigma\in \mathbb{R}^{2\times 2}$.
We choose $\sigma = 0.01\times \mathrm{Id}$ where $\mathrm{Id}$ is the $2\times 2$ identity matrix.
The first class, $\mathbb{P}$, are the set of multivariate Gaussians restricted to $[0,1]^2$ with mean uniformly sampled in $[0,1]^2$ and weighted by $\alpha$ uniformly sampled in $[0.5,1]$.
The second class, $\mathbb{Q}$, are the set of weighted differences between two Gaussian pdf's restricted to $[0,1]^2$ with means $\gamma_1,\gamma_2$ sampled uniformly in $[0,1]^2$.
Note that the second class contains non-positive functions.
See Figure~\ref{fig:App:Syn2:Res} for examples from each class.

We simulate 25 from each set and denote the resulting set of functions by $\mathcal{F}=\{f_i\}_{i=1}^{N}$ where $N=50$.

\paragraph*{Methods.}
Let $(\{f_i\}_{i=1}^N,D_{TL^2_\lambda})$ be a finite dimensional metric space where $D_{TL^2_\lambda}$ is the $N\times N$ matrix containing all pairwise distances is the $TL^2_\lambda$ distance i.e. $D_{TL^2_\lambda}(i,j) = d_{TL^2_\lambda}(f_i,f_j)$.
Similarly for $(\{f_i\}_{i=1}^N,D_{L^2})$ and $(\{f_i\}_{i=1}^N,D_{OT})$ where the optimal transport  distance is defined by $d_{\mathrm{OT}}(f,g) = \sqrt{\mathrm{OT}(f,g)}$ and $\mathrm{OT}$ is given by~\eqref{eq:OTandTLp:Rev:KOT} for $c(x,y) = |x-y|^2_2$.

To apply the optimal transport distance we need to renormalise so that signals are all non-negative and integrate to the same value.
We do this by applying the nonlinear transform $\mathcal{N}(f) = \frac{f-\beta}{\int (f-\beta)}$ where $\beta = \min_{f\in \mathcal{F}} \min_{x\in [0,1]^2} f(x)$.
Neither the $L^2$ nor $TL^2_\lambda$ distances require normalisation.

We use non-metric multidimensional scaling (MDS)~\cite{kruskal64} to represent the graph in $k$ dimensions.
More precisely the aim is to approximate $(\{f_i\}_{i=1}^N,D_{\cdot})$ by a metric space $(\{x_i\}_{i=1}^N,D_{|\cdot|_2})$ embedded in $\mathbb{R}^k$ ($D_{|\cdot|_2}$ is the matrix of pairwise distances using the Euclidean distance, i.e. $D_{|\cdot|_2}(i,j) = |x_i-x_j|_2$).
This is done by minimising the stress $S$ defined by
\[ S_{TL^2_\lambda}(k) = \frac{\sum_{i,j=1}^N \left( |x_i-x_j|_2^2 - F(D_{TL^2_\lambda}(i,j)) \right)^2}{\sum_{i,j=1}^N |x_i-x_j|_2^2} \]
over $\{x_i\}_{i=1}^N \subset\mathbb{R}^k$ and monotonic transformations $F:[0,\infty)\to [0,\infty)$, with $S_{L^2_\lambda}$, $S_{\mathrm{OT}}$ defined analogously.
The classical solution to finding the MDS projection (for Euclidean distances) is to use the $k$ dominant eigenvectors of the matrix of squared distances, after double centring, as coordinates weighted by the square root of the eigenvalue.
More precisely, define  $D^{(2)} = -\frac12 J \left[ |f_i-f_j|^2 \right]_{ij} J$ where $J=\mathrm{Id} - \frac{1}{N} \mathbb{I}$ and $\mathbb{I}$ is the $N\times N$ matrix of ones.
Let $\Lambda_k$ be the matrix with the $k$ largest eigenvalues of $D^{(2)}$ on the diagonal and $E_k$ to be the corresponding matrix of eigenvectors.
Then $X = E_k \Lambda_k^{\frac12}$ is the MDS projection.
Increasing the dimension of the projected space $k$ leads to a better approximation.
In Figure~\ref{fig:App:Syn2:Res} we show the projection in $L^2$, $TL^2$ and the OT distance for $k=2$ as well as the dependence of $k$ on $S$ for each choice of distance.

\paragraph*{Results.}
Our results in Figure~\ref{fig:App:Syn2:Res} show that $TL^2$ is the better distance for this problem.
There is no separation in either $L^2$ or OT whereas $TL^2$ completely separates the data.
It should not therefore be surprising that the 1NN classifier in $TL^2$ outperforms the other distances.
In fact, using 5 fold cross validation (CV) we get 100\% accuracy in $TL^2$, compared to 72\% in $L^2$ and 86\% in OT.
In addition we see that the stress $S_{\rho}$ is much smaller and converges quickly to zero for $TL^2$ which indicates that the $TL^2$ distance is, in this problem, more amenable to a low dimensional representation than either OT or $L^2$.

\subsection{Classification with Real World Data Sets \label{subsec:App:MTS}}

\paragraph*{Objective.}
We evaluate $TL^2_\lambda$ as a distance to classify real world data sets where spatial and intensity information is expected to be important and compare with popular alternative distances.
We choose one dataset which is of the type multivariate time series and a second data set consisting of images.

\paragraph*{Data Sets.}
We use two data sets.
The first is the \emph{AUSLAN}~\cite{kadous02,lichman13} data set which contains 95 classes (corresponding to different words) from a native AUSLAN speaker (Australian Sign Language) using 22 sensors on a CyberGlove (recording position of $x,y,z$ axis, roll, yaw, pitch for left and right hand).
Therefore signals are considered as functions from $\{t_1,t_2,\dots t_N\}$ to $\mathbb{R}^{22}$.
We used the following 25 (out of the 95) classes: alive, all, boy, building, buy, cold, come, computer, cost, crazy, danger, deaf, different, girl, glove, go, God, joke, juice, man, where, which, yes, you and zero.
There are 27 signals in each class which give a total of 675 signals.

We make two pre-processing steps.
The first is to truncate each signal so it is 46 frames in length.
Empirically we find that the signal is constant after the $46^\text{th}$ frame and therefore there is no loss of information in truncating the signal.
The second pre-processing step is to normalise each channel independently.
This is because some channels are orders of magnitude greater than others and would otherwise dominate each choice of distance.

\begin{figure*}[ht]
\begin{figbox}
\centering
\setlength\figureheight{4cm}
\setlength\figurewidth{0.48\columnwidth}
\scriptsize

\begin{subfigure}[t]{0.48\columnwidth}
\centering
\scriptsize
\input{Auslan_Ex.tikz}
\caption{
Example signal from AUSLAN.
\vspace{5pt}
}
\label{subfig:App:MTS:Res:Ex_Auslan}
\end{subfigure}
\hfill
\begin{subfigure}[t]{0.48\columnwidth}
\centering
\scriptsize
%
%
%
%
\begin{tikzpicture}

\begin{axis}[%
width=\figurewidth,
height=\figureheight,
axis on top,
scale only axis,
xmin=0.5,
xmax=84.5,
ymin=0.5,
ymax=56.5,
hide axis
]
\addplot[plot graphics/node/.append style={yscale=-1,anchor=north west}] graphics [xmin=0.5,xmax=84.5,ymin=0.5,ymax=56.5] {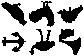};
\end{axis}
\end{tikzpicture}%
\caption{
Example signals from Caltech Silhouettes.
\vspace{5pt}
}
\label{subfig:App:MTS:Res:Ex_CS}
\end{subfigure}

\begin{subfigure}[t]{0.48\columnwidth}
\centering
\scriptsize
\input{Auslan_ClassSep.tikz}
\caption{
A measure of the class identification of each pair of classes for $L^2$ (red squares), DTW (black crosses) and $TL^2$ (blue circles) for AUSLAN.
}
\label{subfig:App:MTS:Res:ClassSep_Auslan}
\end{subfigure}
\hfill
\begin{subfigure}[t]{0.48\columnwidth}
\centering
\scriptsize
%
%
%
%
\begin{tikzpicture}

\begin{axis}[%
width=\figurewidth,
height=\figureheight,
scale only axis,
xmin=-10.91,
xmax=65.45,
ymin=-0.31,
ymax=2.17,
hide axis,
axis background/.style={fill=white!100},
]
\draw [->] (axis cs: 0,0) -- (axis cs: 60,0) node[below] {$ij$};
\draw [->] (axis cs: 0,0) -- (axis cs: 0,2) node[left] {$\kappa_{ij}$};
\draw[-] (axis cs: 0,0) -- ++(2pt,0) node[left,xshift=-2pt] {0};
\draw[-] (axis cs: 0,0.5) -- ++(2pt,0) node[left,xshift=-2pt] {0.5};
\draw[-] (axis cs: 0,1) -- ++(2pt,0) node[left,xshift=-2pt] {1};
\draw[-] (axis cs: 0,1.5) -- ++(2pt,0) node[left,xshift=-2pt] {1.5};
\draw[-] (axis cs: 0,0) -- ++(0,2pt);
\draw[-] (axis cs: 10,0) -- ++(0,2pt);
\draw[-] (axis cs: 20,0) -- ++(0,2pt);
\draw[-] (axis cs: 30,0) -- ++(0,2pt);
\draw[-] (axis cs: 40,0) -- ++(0,2pt);
\draw[-] (axis cs: 50,0) -- ++(0,2pt);
\draw[-] (axis cs: 60,0) -- ++(0,2pt);
\addplot [
color=red,
mark size=0.9pt,
only marks,
mark=square*,
mark options={solid},
forget plot
]
table[row sep=crcr]{
1 0.888290068697967\\
2 0.926392796322318\\
3 0.950233700621897\\
4 0.952371461930362\\
5 0.954504435401516\\
6 0.960874947278021\\
7 0.975577713059844\\
8 0.98388065182025\\
9 1.00027979217845\\
10 1.00433773903212\\
11 1.00837935592104\\
12 1.01039410189763\\
13 1.01039410189763\\
14 1.01841322848669\\
15 1.01841322848669\\
16 1.02040816326531\\
17 1.02040816326531\\
18 1.03229681481114\\
19 1.03426497067773\\
20 1.04405009845312\\
21 1.04793856665529\\
22 1.04987740007362\\
23 1.04987740007362\\
24 1.0556725355387\\
25 1.0556725355387\\
26 1.06526103880676\\
27 1.06526103880676\\
28 1.07476400168066\\
29 1.07665452909019\\
30 1.07665452909019\\
31 1.07854174268417\\
32 1.08042565982764\\
33 1.08230629773451\\
34 1.08979639608642\\
35 1.09538035952281\\
36 1.09908724104682\\
37 1.10278166235909\\
38 1.10278166235909\\
39 1.1101336215261\\
40 1.13728026148769\\
41 1.14263219816128\\
42 1.15678298132762\\
43 1.16204509938065\\
44 1.16728349602826\\
45 1.17249848921239\\
46 1.17941594444559\\
47 1.18971671595521\\
48 1.19313054425189\\
49 1.20668928114952\\
50 1.21173469387755\\
51 1.21173469387755\\
52 1.2167591854808\\
53 1.26590773222444\\
54 1.29450147513931\\
55 1.32247712712549\\
};
\addplot [
color=black,
mark size=1.5pt,
only marks,
mark=x,
mark options={solid},
forget plot
]
table[row sep=crcr]{
1 1.43746383731367\\
2 1.2357677612778\\
3 1.36433440326257\\
4 1.45794024365073\\
5 1.10861088261146\\
6 1.08209899403583\\
7 1.48366155782699\\
8 1.36025289250241\\
9 0.933596303421155\\
10 1.17029493053546\\
11 1.02924506438123\\
12 0.926804643949312\\
13 1.12704212888674\\
14 0.90789503811939\\
15 1.91688315296648\\
16 1.51339888522785\\
17 1.272338950373\\
18 1.50222476368948\\
19 1.06004560759114\\
20 1.01997358551925\\
21 1.02062977975456\\
22 0.871030128987932\\
23 1.4562459888471\\
24 1.27583115488898\\
25 1.36498171046019\\
26 1.62023394373531\\
27 1.17920974248511\\
28 0.929551849014419\\
29 0.962968732074808\\
30 1.35599814642513\\
31 1.08838121303644\\
32 1.19285859096259\\
33 1.85797586783643\\
34 1.21495937471099\\
35 1.44608597977731\\
36 1.09575162695795\\
37 1.63836465288625\\
38 1.98132494267482\\
39 1.2768504226163\\
40 1.54099698970845\\
41 1.34329635775309\\
42 1.18125079141695\\
43 1.90326209298907\\
44 1.14543555159579\\
45 0.966858358835207\\
46 1.98214107683514\\
47 1.22666551438672\\
48 0.972374614549307\\
49 1.10684159722156\\
50 1.44941954808849\\
51 1.86406885691415\\
52 1.08100567694628\\
53 1.67059172776064\\
54 1.00059696970572\\
55 1.36692176827466\\
};
\addplot [
color=blue,
mark size=0.9pt,
only marks,
mark=*,
mark options={solid},
forget plot
]
table[row sep=crcr]{
1 1.23093299166781\\
2 1.04064764297551\\
3 1.3537136686845\\
4 1.23841594807715\\
5 1.46890091002617\\
6 1.32010020490534\\
7 1.20733870073386\\
8 1.2302996442096\\
9 1.018167123403\\
10 1.4428251288122\\
11 1.06575039917843\\
12 1.44817789777866\\
13 1.06995532007575\\
14 1.04612356032224\\
15 1.4200609973829\\
16 1.33103677131792\\
17 1.37123636564948\\
18 1.33646434434124\\
19 1.07980064113739\\
20 1.07128119720208\\
21 1.53933816928989\\
22 1.58581407998002\\
23 1.39954496001691\\
24 1.14400669293979\\
25 1.17477856973252\\
26 1.17001063035946\\
27 1.56787412773763\\
28 1.14515712536533\\
29 1.1181594251678\\
30 1.18130359048285\\
31 1.17170152006588\\
32 1.27947772390341\\
33 1.24119335254237\\
34 1.17015458436492\\
35 1.13002421076529\\
36 1.2060716981\\
37 1.23581352221905\\
38 1.54220331510709\\
39 1.15143070758391\\
40 1.49072925915969\\
41 1.42038375752113\\
42 1.69895075245128\\
43 1.48526103144938\\
44 1.29272814014766\\
45 1.26137159616866\\
46 1.47102309520329\\
47 1.35614605444132\\
48 1.39057254030725\\
49 1.32317985657606\\
50 1.49409229823176\\
51 1.37354810242164\\
52 1.2929709136517\\
53 1.60280556159691\\
54 1.63377143350662\\
55 1.65902583682239\\
};
\end{axis}
\end{tikzpicture}%
\caption{
A measure of the class identification of each pair of classes for $L^2$ (red squares), OT (black crosses) and $TL^2$ (blue circles) for Caltech Silhouettes.
}
\label{subfig:App:MTS:Res:ClassSep_CS}
\end{subfigure}

\caption{
Example signals and results for the data sets described in Section~\ref{subsec:App:MTS}.
}
\label{fig:App:MTS:Res}
\end{figbox}
\end{figure*}

The second data set we use is a subset of the $28\times 28$ Caltech Silhouettes database~\cite{marlin10}.
This data set was derived from the Caltech 101 data set~\cite{feifei04}, which consists of images from 101 categories, by finding and filling in the outline for the object of focus in each image.
See Figure~\ref{subfig:App:MTS:Res:Ex_CS} for examples.
The subset we uses consists of the following 11 images: anchor, barrel, crocodile head, dollar bill, emu, gramophone, pigeon, pyramid, rhino, rooster and stegosaurus.
The number of images in each class varied from 42 to 59.
There were 565 images in total.
 
\paragraph*{Methods.}
For the multivariate time series we compare the performance of a 1NN classifier using the $L^2$ and $TL_\lambda^2$ distances as well as the state-of-the-art method dynamic time warping~\cite{gorecki15}.
There are three common variations of dynamic time warping.
One can apply dynamic time warping directly to the signals $f$ and $g$ (denoted by DTW), to the derivative $f'$ of the signals (denoted by DDTW) and to a weighted average of DTW and DDTW (denoted by WDTW).
We define
\begin{align*}
d_{DDTW}(f,g) & = d_{DTW}(f',g') \\
d_{WDTW}(f,g) & = \alpha d_{DTW}(f,g) + (1-\alpha) d_{DDTW}(f,g).
\end{align*}
The parameter $\alpha$ is chosen by 5-fold 2nd depth cross validation.

One can define the analogous distances for $L^2$ and $TL^2$ by
\begin{align*}
d_{DL^2}(f,g) & = d_{L^2}(f',g') \\
d_{DTL^2_\lambda}(f,g) & = d_{TL^2_\lambda}(f',g') \\
d_{WL^2}(f,g) & = \alpha d_{L^2}(f,g) + (1-\alpha) d_{DL^2}(f,g) \\
d_{WTL^2_\lambda}(f,g) & = \alpha d_{TL^2_\lambda}(f,g) + (1-\alpha) d_{DTL^2_\lambda}(f,g).
\end{align*}
We do not have to choose the same value of $\lambda$ in $TL^2$ and $DTL^2$ however considering that signals are normalised, we will use the same value.
Note that $DL^2$, DTW, DDTW, WDTW and $DTL^2_\lambda$ are \emph{not} metrics.

We remark that an alternative method for including derivatives in the $TL^p$ distance would be to extend the signal to include the derivative.
We briefly assume that $f$ is defined over a continuous domain.
Let $f:\mathbb{R}\to \mathbb{R}$, and $\tilde{f} = \left(f,\frac{\mathrm{d}f}{\mathrm{d} x}\right)$ then we define
\[ d_{TW^{1,p}_\lambda}(f,g) = d_{TL^p_\lambda}(\tilde{f},\tilde{g}). \]
We take our notation $TW^{k,p}_\lambda$ from the Sobolev space notation where $W^{k,p}$ is the Sobolev space with $k$ weak derivatives integrable in $L^p$.
There is no reason to limit this to one derivative, and we may define $\tilde{f} = \left(f,\frac{\mathrm{d}f}{\mathrm{d} x},\dots,\frac{\mathrm{d}^kf}{\mathrm{d} x^k}\right)$ and
\[ d_{TW^{k,p}_\lambda}(f,g) = d_{TL^p_\lambda}(\tilde{f},\tilde{g}). \]
When the signals are discrete one should use a discrete approximation of the derivative.
In order to be consistent with previous extensions of dynamic time warping we do not develop this approach here.

Dynamic time warping is only defined on time series so we are not able to apply it to the Caltech Silhouettes database.
Instead we use the optimal transport distance (with $p=2$).
To apply the optimal transport distance each image $f\in \mathbb{R}^2\to \{0,1\}$ is normalised by $\hat{f}(x)=\frac{f(x)}{\int_{[0,1]^2} f(y) \, \mathrm{d} y}$.
There is no normalisation for either $L^2$ or $TL^2$.
We find the 1NN classifier using $TL^2$, $L^2$ and OT distances.

We will use $\lambda=1$ in AUSLAN and $\lambda=0.1$ in Caltech Silhouettes for the $TL^2$ based distances.
The underlying measure $\mu$ is chosen to be the uniform measure defined on $[0,1]$ or $[0,1]^2$.

\paragraph*{Results.}
We considered two methods for comparing the performance of each distance.
The first is the 1NN classification accuracy in each distance.
We use the 1NN classification accuracy as a measure as to how well each distance captures the underlying geometry.
A higher accuracy implies closest neighbours are more likely to belong to the same class.

The results are given in Table~\ref{tab:App:MTS:Res} where we report error rates using 5-fold cross-validation.
In terms of the 1NN classifier for the AUSLAN data set we see that $TL^2$ is significantly better than $L^2$ and is a modest improvement over dynamic time warping.
And for the Caltech Silhouettes dataset OT performs poorly with $TL^2$ the best performer.

In the same spirit as Section~\ref{subsec:App:Syn1} we define the performance metric $\kappa_{ij}(\rho)$ as the ratio of distance between class $i$ and class $j$ and the maximum class coverage radius of class $i$ and class $j$.
For the distance between classes we use the Hausdorff distance (see Section~\ref{subsec:App:Syn1}) and for the class coverage radius we use the minimum radius $r$ such that connecting any two data points in class $i$ closer than $r$ defines a connected graph.
We plot the results in Figures~\ref{subfig:App:MTS:Res:ClassSep_Auslan} and~\ref{subfig:App:MTS:Res:ClassSep_CS}.
The $x$ axis represents pairs of classes where for visual clarity we have ordered the pairs so that the $\kappa(L^2)$ is increasing.
A large value of $\kappa_{ij}$ indicates that it is easier to identify class $i$ from class $j$ whereas a small value indicates that identifying the two classes is a difficult problem.

For AUSLAN we see that $TL^2$ has, for the majority of pairs of classes, a larger value of $\kappa_{ij}$ than $L^2$ and $DTW$ and therefore better represents the class structure.
For the Caltech Silhouettes dataset $L^2$ has the worst separation even though it outperformed OT in the 1NN test.
The $TL^2$ distance is much more consistent than OT, we can see that although between some classes OT is the best distance with other classes OT does extremely poorly (worse than $L^2$).
On the other hand $TL^2$ is better than $L^2$ for every pair of classes.

\begin{table*}[ht]
\centering
\begin{tabular}{cccccccccc}
\toprule
Dataset & $L^2$ & $DL^2$ & $WL^2$ & DTW & DDTW & WDTW & $TL^2_\lambda$ & $DTL^2_\lambda$ & $WTL^2_\lambda$ \\
\toprule
AUSLAN &
10.4 & 14.7 & 9.8 & 
9.6 & 3.1 & 2.7 & 
9.5 & 2.4 & 2.7 \\ 
\bottomrule
\end{tabular}

\begin{tabular}{cccc}
\toprule
Dataset & $L^2$ & OT & $TL^2_\lambda$ \\
\toprule
Caltech Silhouettes &
12.8 & 
19.6 & 
11.0 \\ 
\bottomrule
\end{tabular}
\vspace{5pt}
\caption{
Error rates (\%) for 1NN classification.
}
\label{tab:App:MTS:Res}
\end{table*}

\subsection{Histogram Specification and Colour Transfer with \texorpdfstring{$TL^p$}{TLp} \label{subsec:App:Colour}}

\paragraph*{Histogram specification and colour transfer.}
Histogram specification concerns the problem of matching one histogram onto another.
For a function $f$ on a discrete domain $X$ the histogram is given by $f_{\#}\mu$ where $\mu$ is the uniform discrete measure supported on $N$ points.
We do not make any assumption on the dimension of the codomain of $f$ (so that $f$ may be multivalued and the histogram may be multidimensional).
This coincides with the definition given in the introduction to the section, that is
\[ f_{\#}\mu(y) = \frac{1}{N}\#\left\{ x\in X \, : \, f(x) = y \right\}. \]
Given two functions $f:X\to \mathbb{R}^m$ and $g:Y\to \mathbb{R}^m$, with histograms $\varphi$ and $\psi$ respectively, histogram specification is the problem of finding a map $T:X\to Y$ such that $\psi = T_{\#}\varphi$.

The colour transfer problem is the problem of colouring one image $f$ with the palette of an exemplar image $g$.
A common method used to solve this problem is to use histogram specification where $T$ is the minimizer to Monge's optimal transport problem~\eqref{eq:OTandTLp:Rev:MOT} between $\varphi$ and $\psi$~\cite{chizat16,ferradans13,morovic03,rabin14,rabin15}.
Let our colour space be denoted by $\mathcal{C}$ where for example if the colour space is 8 bit RGB then $\mathcal{C} = \{0,1,\dots,255\}^3$.
The colour histogram then defines a measure over $\mathcal{C}$.
If we consider two such histograms $\varphi$ and $\psi$ corresponding to images $f:X\to \mathcal{C}$ and $g:Y\to \mathcal{C}$ respectively then a histogram specification is a map $T:\mathcal{C} \to \mathcal{C}$ that satisfies $\psi = T_{\#} \varphi$.
The recoloured image $\hat{f} = g \circ T$ has the same colour histogram as $g$.
The solution $\hat{f}$ is a recolouring of $f$ using the palette of $g$.

If we consider grayscale images then $\mathcal{C} = [0,1]$ and the optimal transport map (assuming it exists) is a monotonically increasing function.
In particular this implies that if pixel $x$ is lighter than pixel $y$ (i.e. $f(x) > f(y)$) then in the recoloured image $\hat{f} = T\circ f$ pixel $x$ is still lighter than pixel $y$.
In this sense the OT solution preserves intensity ordering.
But note that no spatial information is used to define $T$; only the difference in intensity between pixels is used and not the distance between pixels.

\paragraph*{Spatially correlated histogram specification.}
Let $\varphi$ and $\psi$ be the histograms corresponding to images $f:X\to \mathbb{R}^m$ and $g:Y\to \mathbb{R}^m$ respectively.
If we recall Proposition~\ref{prop:Frame:lambdatoInfty} then $\lim_{\lambda\to \infty} d_{TL_\lambda^p}((f,\mu),(g,\nu)) = d_{\mathrm{OT}}(f_{\#}\mu,g_{\#} \nu)$ (where $\mu$ and $\nu$ are the discrete uniform measures over the sets $X$ and $Y$).
For $\lambda<\infty$ the $TL^p_\lambda$ distance includes spatial \emph{and} intensity information.
Hence the $TL_\lambda^p$ distance provides a generalization of OT induced histogram specification.

Analogously to the OT induced histogram specification method we define the spatially correlated histogram specification to be histogram specification using the map $T:X\to Y$ which is a minimizer to Monge's formulation of the $TL_\lambda^p$ distance~\eqref{eq:Frame:TLpMonge}.
When the images are of the same size then, by Proposition~\ref{prop:Frame:Def:ExistTransMap} such a map exists.
The recoloured image $\hat{f}$ of $f$ is given by $\hat{f} = g\circ T$.
Furthermore when the images are of the same size the map $T$ is a rearrangement of the pixels in $X$ and therefore the histograms are invariant under $T$.
In particular the histogram of $\hat{f}$ is the same as the histogram of $g$.

Although we propose the spatially correlated histogram specification as a method to incorporate spatial structure we now point out its value as a numerically efficient approximation to OT induced histogram specification for colour images.
Motivated by Proposition~\ref{prop:Frame:lambdatoInfty} one expects that for large $\lambda$ the $TL^p_\lambda$ map is approximately the OT map between colour histograms.
The OT problem is in the $\mathcal{C}$ space which, for colour images is 3 dimensional.
However, the $TL_\lambda^p$ problem is in the domain of the images, which is typically 2 dimensional.
Hence one can use $TL^p_\lambda$ to approximate OT induced histogram specification in a lower dimensional space.

We briefly remark that histogram specification methods often include additional regularization terms.
Such choices of regularization on the transport map include penalizing the gradients~\cite{ferradans13,rabin14,rabin15}, sparsity~\cite{rabin15} and average transport~\cite{papadakis12}.
One could apply any of the above regularizations to spatially correlated histogram specification.

\begin{figure*}
\begin{figbox}
\centering
\setlength\figurewidth{0.23\columnwidth}
\setlength\figureheight{0.23\columnwidth}
\scriptsize

\begin{subfigure}[t]{0.24\columnwidth}
\centering
\scriptsize
%
%
%
%
\begin{tikzpicture}

\begin{axis}[%
width=\figurewidth,
height=\figureheight,
axis on top,
scale only axis,
xmin=0.5,
xmax=128.5,
ymin=0.5,
ymax=128.5,
hide axis,
]
\addplot[plot graphics/node/.append style={yscale=-1,anchor=north west}] graphics [xmin=0.5,xmax=128.5,ymin=0.5,ymax=128.5] {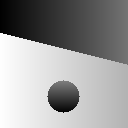};
\end{axis}
\end{tikzpicture}%
\caption{
Exemplar image.
}
\label{subfig:App:Colour:Grayscale:Image1}
\end{subfigure}
\hfill
\begin{subfigure}[t]{0.24\columnwidth}
\centering
\scriptsize
%
%
%
%
\begin{tikzpicture}

\begin{axis}[%
width=\figurewidth,
height=\figureheight,
axis on top,
scale only axis,
xmin=0.5,
xmax=128.5,
ymin=0.5,
ymax=128.5,
hide axis,
]
\addplot[plot graphics/node/.append style={yscale=-1,anchor=north west}] graphics [xmin=0.5,xmax=128.5,ymin=0.5,ymax=128.5] {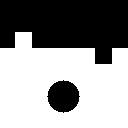};
\end{axis}
\end{tikzpicture}%
\caption{
Original image to be shaded.
}
\label{subfig:App:Colour:Grayscale:Image2}
\end{subfigure}
\hfill
\begin{subfigure}[t]{0.24\columnwidth}
\centering
\scriptsize
%
%
%
%
\begin{tikzpicture}

\begin{axis}[%
width=\figurewidth,
height=\figureheight,
axis on top,
scale only axis,
xmin=0.5,
xmax=128.5,
ymin=0.5,
ymax=128.5,
hide axis,
]
\addplot[plot graphics/node/.append style={yscale=-1,anchor=north west}] graphics [xmin=0.5,xmax=128.5,ymin=0.5,ymax=128.5] {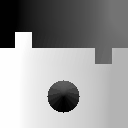};
\end{axis}
\end{tikzpicture}%
\caption{
The $TL^p_\lambda$ solution.
}
\label{subfig:App:Colour:Grayscale:Reslambda2}
\end{subfigure}
\hfill
\begin{subfigure}[t]{0.24\columnwidth}
\centering
\scriptsize
\definecolor{mycolor1}{rgb}{0.2,0.2,0.2}

\begin{tikzpicture}
\begin{axis}[
width=\figurewidth,
height=0.5\figureheight,
scale only axis,
xmin=-0.1,
xmax=1.2,
ymin=-0.12,
ymax=0.66,
hide axis,
name=plot1,
axis background/.style={fill=white!100},
]
\draw [->] (axis cs: 0,0) -- (axis cs: 1.1,0);
\draw [->] (axis cs: 0,0) -- (axis cs: 0,0.6);
\addplot[ybar,bar width=1.3pt,draw=mycolor1,fill=mycolor1,fill] table [row sep=crcr] {
0.01	0.42364501953125\\
0.03	0\\
0.05	0\\
0.07	0\\
0.09	0\\
0.11	0\\
0.13	0\\
0.15	0\\
0.17	0\\
0.19	0\\
0.21	0\\
0.23	0\\
0.25	0\\
0.27	0\\
0.29	0\\
0.31	0\\
0.33	0\\
0.35	0\\
0.37	0\\
0.39	0\\
0.41	0\\
0.43	0\\
0.45	0\\
0.47	0\\
0.49	0\\
0.51	0\\
0.53	0\\
0.55	0\\
0.57	0\\
0.59	0\\
0.61	0\\
0.63	0\\
0.65	0\\
0.67	0\\
0.69	0\\
0.71	0\\
0.73	0\\
0.75	0\\
0.77	0\\
0.79	0\\
0.81	0\\
0.83	0\\
0.85	0\\
0.87	0\\
0.89	0\\
0.91	0\\
0.93	0\\
0.95	0\\
0.97	0\\
0.99	0.57635498046875\\
};
\end{axis}

\begin{axis}[
width=\figurewidth,
height=0.5\figureheight,
scale only axis,
xmin=-0.1,
xmax=1.2,
ymin=-0.012,
ymax=0.066,
hide axis,
at=(plot1.above north west),
axis background/.style={fill=white!100},
]
\draw [->] (axis cs: 0,0) -- (axis cs: 1.1,0);
\draw [->] (axis cs: 0,0) -- (axis cs: 0,0.06);
\addplot[ybar,bar width=1.3pt,draw=mycolor1,fill=mycolor1,fill] table [row sep=crcr] {
0.01	0.01092529296875\\
0.03	0.01361083984375\\
0.05	0.01446533203125\\
0.07	0.01416015625\\
0.09	0.01580810546875\\
0.11	0.0164794921875\\
0.13	0.014404296875\\
0.15	0.0194091796875\\
0.17	0.01824951171875\\
0.19	0.0159912109375\\
0.21	0.02117919921875\\
0.23	0.01983642578125\\
0.25	0.01739501953125\\
0.27	0.02081298828125\\
0.29	0.0233154296875\\
0.31	0.02191162109375\\
0.33	0.01904296875\\
0.35	0.02459716796875\\
0.37	0.02325439453125\\
0.39	0.02008056640625\\
0.41	0.02410888671875\\
0.43	0.02587890625\\
0.45	0.012939453125\\
0.47	0.00091552734375\\
0.49	0.00067138671875\\
0.51	6.103515625e-005\\
0.53	0\\
0.55	0\\
0.57	0\\
0.59	0\\
0.61	0\\
0.63	0\\
0.65	0\\
0.67	0\\
0.69	0\\
0.71	0.03533935546875\\
0.73	0.03662109375\\
0.75	0.0335693359375\\
0.77	0.0389404296875\\
0.79	0.03564453125\\
0.81	0.0384521484375\\
0.83	0.0247802734375\\
0.85	0.02642822265625\\
0.87	0.02685546875\\
0.89	0.04315185546875\\
0.91	0.0418701171875\\
0.93	0.04827880859375\\
0.95	0.0439453125\\
0.97	0.05059814453125\\
0.99	0.0460205078125\\
};
\end{axis}
\end{tikzpicture}
\caption{
The grayscale histograms of the exemplar image (top) and the original image (bottom).
}
\label{subfig:App:Colour:Grayscale:Hist}
\end{subfigure}

\caption{
Spatially correlated histogram specification of synthetic grayscale images.
}
\label{fig:App:Colour:Grayscale}
\end{figbox}
\end{figure*}

\paragraph*{Examples.}
First, let us consider the grayscale images in Figure~\ref{fig:App:Colour:Grayscale}.
The objective is to combine the shading of the first image with the geometry of the second image.
We are motivated by the scenario where one wishes to combine information about a scene obtained by two different measurements: one where intensities (dynamical range) are well resolved, but the spatial resolution (geometry) is not well captured, and another where dynamical range is poorly captured, but the geometry is well resolved. We furthermore allow that the scenes captured may be somewhat different.
The desire is to combine the images to obtain a single image with both good geometry and intensity.
The solution we propose is to use spatially correlated histogram specification to re-shade the image with low quality intensity.

The result, as given in Figure~\ref{fig:App:Colour:Grayscale}, produces what we consider to be the desired output.
The shading has been transferred and the geometry has not been lost.
One is not able to apply histogram specification (induced by the OT map) due to the lack of existence of an optimal transport map from the histogram of the original image $\varphi$ to the histogram exemplar image $\psi$.
This is due to the histogram of the original image being a sum of two delta masses as in Figure~\ref{subfig:App:Colour:Grayscale:Hist}.

\begin{figure*}[ht]
\begin{figbox}
\centering
\setlength\figureheight{0.25\columnwidth}
\setlength\figurewidth{0.3\columnwidth}
\scriptsize

\begin{subfigure}[t]{0.45\columnwidth}
\centering
\scriptsize
%
%
%
%
\begin{tikzpicture}

\begin{axis}[%
width=\figurewidth,
height=\figureheight,
axis on top,
scale only axis,
xmin=0.5,
xmax=180.5,
ymin=0.5,
ymax=150.5,
hide axis,
]
\addplot[plot graphics/node/.append style={yscale=-1,anchor=north west}] graphics [xmin=0.5,xmax=180.5,ymin=0.5,ymax=150.5] {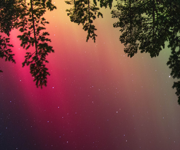};
\end{axis}
\end{tikzpicture}%
\caption{
Exemplar image.
\vspace{5pt}
}
\label{subfig:App:Colour:NL:Image1}
\end{subfigure}
\hfill
\begin{subfigure}[t]{0.45\columnwidth}
\centering
\scriptsize
%
%
%
%
\begin{tikzpicture}

\begin{axis}[%
width=\figurewidth,
height=\figureheight,
axis on top,
scale only axis,
xmin=0.5,
xmax=180.5,
ymin=0.5,
ymax=150.5,
hide axis,
]
\addplot[plot graphics/node/.append style={yscale=-1,anchor=north west}] graphics [xmin=0.5,xmax=180.5,ymin=0.5,ymax=150.5] {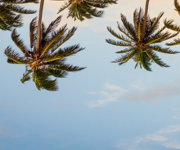};
\end{axis}
\end{tikzpicture}%
\caption{
Original image to be coloured.
\vspace{5pt}
}
\label{subfig:App:Colour:NL:Image2}
\end{subfigure}

\begin{subfigure}[t]{0.45\columnwidth}
\centering
\scriptsize
%
%
%
%
\begin{tikzpicture}

\begin{axis}[%
width=\figurewidth,
height=\figureheight,
axis on top,
scale only axis,
xmin=0.5,
xmax=180.5,
ymin=0.5,
ymax=150.5,
hide axis,
]
\addplot[plot graphics/node/.append style={yscale=-1,anchor=north west}] graphics [xmin=0.5,xmax=180.5,ymin=0.5,ymax=150.5] {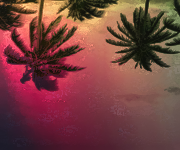};
\end{axis}
\end{tikzpicture}%
\caption{
$TL^p_\lambda$ solution for $\lambda = 0.1$.
\vspace{5pt}
}
\label{subfig:App:Colour:NL:Reslambda0p1}
\end{subfigure}
\hfill
\begin{subfigure}[t]{0.45\columnwidth}
\centering
\scriptsize
%
%
%
%
\begin{tikzpicture}

\begin{axis}[%
width=\figurewidth,
height=\figureheight,
axis on top,
scale only axis,
xmin=0.5,
xmax=180.5,
ymin=0.5,
ymax=150.5,
hide axis,
]
\addplot[plot graphics/node/.append style={yscale=-1,anchor=north west}] graphics [xmin=0.5,xmax=180.5,ymin=0.5,ymax=150.5] {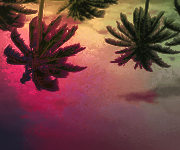};
\end{axis}
\end{tikzpicture}%
\caption{
$TL^p_\lambda$ solution for $\lambda = 1$.
\vspace{5pt}
}
\label{subfig:App:Colour:NL:Reslambda1}
\end{subfigure}

\begin{subfigure}[t]{0.45\columnwidth}
\centering
\scriptsize
%
%
%
%
\begin{tikzpicture}

\begin{axis}[%
width=\figurewidth,
height=\figureheight,
axis on top,
scale only axis,
xmin=0.5,
xmax=180.5,
ymin=0.5,
ymax=150.5,
hide axis,
]
\addplot[plot graphics/node/.append style={yscale=-1,anchor=north west}] graphics [xmin=0.5,xmax=180.5,ymin=0.5,ymax=150.5] {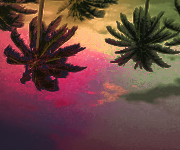};
\end{axis}
\end{tikzpicture}%
\caption{
$TL^p_\lambda$ solution for $\lambda = 10$.
}
\label{subfig:App:Colour:NL:Reslambda10}
\end{subfigure}
\hfill
\begin{subfigure}[t]{0.45\columnwidth}
\centering
\scriptsize
%
%
%
%
\begin{tikzpicture}

\begin{axis}[%
width=\figurewidth,
height=\figureheight,
axis on top,
scale only axis,
xmin=0.5,
xmax=180.5,
ymin=0.5,
ymax=150.5,
hide axis,
]
\addplot[plot graphics/node/.append style={yscale=-1,anchor=north west}] graphics [xmin=0.5,xmax=180.5,ymin=0.5,ymax=150.5] {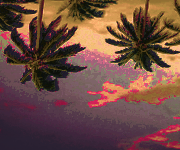};
\end{axis}
\end{tikzpicture}%
\caption{
OT colour transfer solution (no spatial information).
}
\label{subfig:App:Colour:NL:ResOT}
\end{subfigure}

\caption{
Spatially correlated histogram specification of real colour images.
}
\label{fig:App:Colour:NL}
\end{figbox}
\end{figure*}

As a more challenging example we consider the colour images in Figure~\ref{fig:App:Colour:NL}.
The exemplar image contains a few trees with the northern lights in the background, whilst the other image has a few trees with a mostly clear sky in the background.
The challenge is to recreate the northern lights in the second image.

As one would expect, in Figure~\ref{subfig:App:Colour:NL:ResOT} we see that the histogram specification induced by OT loses the spatial structure.
Indeed, it is hard to recognise the northern lights.
The spatially correlated histogram specification solution does a much better job at preserving the ordering locally.
As $\lambda$ increases it becomes cheaper to match pixels that are further apart and therefore, for large $\lambda$, the matching does not preserve the local structure in the exemplar image.

\section{Conclusions \label{sec:conc}}

In this paper we have developed and applied a distance that directly accounts for the intensity of the signal within a Lagrangian framework.
This differs from optimal transport that does not directly measure intensity and the $L^p$ distance which measures intensity only.
Through applications we have shown the potential of this distance in signal analysis.

The distance is widely applicable, unlike optimal transport the distance does not require treating signals as measures.
Treating a signal as a measure implies the following constraints: non-negative signals, conservation of mass and single channelled signals.
None of these assumptions are necessary for the $TL^p_\lambda$ distance.
Furthermore the framework is general enough to include discrete and continuous signals as well as allowing practitioners to emphasise features which in many cases should allow for a better representation of data sets, for example one could include derivatives.

Efficient existing methods, such as entropy regularized or multi-scale linear programming, for optimal transport are applicable to the $TL^p_\lambda$ distance.
In fact any numerical method for optimal transport that can cope with arbitrary cost functions is immediately available.

Via the representation as an optimal transport distance between measures supported on graphs we expect many other results for OT to carry through to $TL^p$.
For example, one could extend the LOT method~\cite{wang12} for signal representation and analysis to the $TL^p$ framework.
This would allow pairwise distances of a data set to be computed with numerical cost that is linear in number of images.
We leave the development for future work.

The applications we considered were classification and histogram specification in the context of colour transfer.
For classification we chose data sets with a Lagrangian nature but were either multi-channelled (so that optimal transport distances are not available) or non-positive (in which case one has to rescale in order to apply optimal transport).
We showed the $TL^p_\lambda$ distance better represented the underlying geometry.
For the colour transfer problem we defined a spatially correlated histogram specification method which produced more visually appealing results when combining the colour of one image with the geometry of another.

Although the main motivation was to develop a distance which better represents Lagrangian data sets we also note that the $TL^p_\lambda$ distance provides a numerically efficient approximation for the optimal transport induced histogram specification method by, for 2-dimensional images colour images, reducing the effective dimension of the problem from three for optimal transport to two for $TL^p_\lambda$.
We also observe that the effective dimension of multi-channelled time signals is one.
In particular the effective dimension is independent of the number of channels.

The applications we have considered are for demonstration on the performance of $TL^p_\lambda$.
A next step would be to consider a more detailed study of a specific problem.
For example in the colour transfer application we could have considered regularization terms which would have improved the performance.
It was not the aim to propose a state-of-the-art method for each application, indeed each application would constitute a paper within its own right.

\subsection*{Acknowledgements}

Authors gratefully acknowledge funding from the NSF (CCF 1421502) and the NIH (GM090033, CA188938) in contributing to a portion of this work. DS also acknowledges funding by NSF ( DMS-1516677).
The authors are also grateful to the Center for Nonlinear Analysis of CMU for its support. 

\appendix

\section{Performance of \texorpdfstring{$TL^p_\lambda$}{TLp} in Classification Problems with Simple and Oscillatory Signals \label{sec:app:ClassDetails}}

We compare the performance of $TL_\lambda^2$, $L^2$ and OT distances with respect to classification/clustering for the three classes $\{\mathcal{C}_i\}_{i=1,2,3}$ of signals defined in Figure~\ref{fig:App:Syn1:Classes}.
We test how each distance performs by finding the smallest number of data points such that the classes $\mathcal{C}_i^N=\{f_i\}_{i=1}^N\subset \mathcal{C}_i$ are separable.
For sufficiently large $N$ the approximation $d_{H,\rho}(\mathcal{C}_i^N,\mathcal{C}_j^N) \approx d_{H,\rho}(\mathcal{C}_i,\mathcal{C}_j)$ is used to simplify the computation.
Similarly, as a proxy for $\mathbb{E}R_\rho(\mathcal{C}_i^N)$ we use $R_\rho(\hat{\mathcal{C}}^N)$ where 
\begin{align*}
\hat{\mathcal{C}}_i^N = \Bigg\{ f_\ell \, : \, & \ell = \ell_{\min}^i + \frac{n-1}{N-1} \left(\ell_{\min}^i-\ell_{\max}^i \right) , \\
 & \quad n\in \{1,2,\dots, N\} \Bigg\}
\end{align*} 
is the uniform sample from class $\mathcal{C}_i$ (recall that class $\mathcal{C}_i$ is parameterized by $\ell\in [\ell_{\min}^i,\ell_{\max}^i]$ and with an abuse of notation we use the subscript of $f_\ell$ to denote the dependence of $\ell$).

It follows that the class separation distances and class coverage radius are approximated by
\begin{align*}
d_{H,L^2}^2(\mathcal{C}_1^N,\mathcal{C}_2^N) & \approx \frac{\alpha}{2}
& R^2_{L^2}(\mathcal{C}_1^N) & \approx \frac{2}{N} \\
d_{H,L^2}^2(\mathcal{C}_1^N,\mathcal{C}_3^N) & \approx \frac{3\alpha}{4}
& R^2_{L^2}(\mathcal{C}_2^N) & \approx \frac{1}{N} \\
d_{H,L^2}^2(\mathcal{C}_2^N,\mathcal{C}_3^N) & \approx \frac{\alpha}{4}
& R^2_{L^2}(\mathcal{C}_3^N) & \approx \frac{2\alpha}{N\gamma} \\
d_{H,\mathrm{OT}}^2(\mathcal{C}_1^N,\mathcal{C}_2^N) & \approx \frac{\beta^2\alpha}{4}
& R^2_{\mathrm{OT}}(\mathcal{C}_1^N) & \approx \frac{\alpha}{N^2} \\
d_{H,\mathrm{OT}}^2(\mathcal{C}_1^N,\mathcal{C}_3^N) & \approx \frac{\beta^2\alpha}{4}
& R^2_{\mathrm{OT}}(\mathcal{C}_2^N) & \approx \frac{\alpha}{N^2} \\
d_{H,\mathrm{OT}}^2(\mathcal{C}_2^N,\mathcal{C}_3^N) & \approx \frac{\alpha\gamma^2}{8}
& R^2_{\mathrm{OT}}(\mathcal{C}_3^N) & \approx \frac{\alpha}{N^2} \\
d_{H,TL^2_\lambda}^2(\mathcal{C}_1^N,\mathcal{C}_2^N) & \approx \frac{\alpha}{2}
& R^2_{TL^2_\lambda}(\mathcal{C}_1^N) & \approx \frac{\alpha^2}{N} \\
d_{H,TL^2_\lambda}^2(\mathcal{C}_1^N,\mathcal{C}_3^N) & \approx \frac{3\alpha}{4}
& R^2_{TL^2_\lambda}(\mathcal{C}_2^N) & \approx \frac{4\alpha^2}{N}
\end{align*}
\begin{align*}
d_{H,TL^2_\lambda}^2(\mathcal{C}_2^N,\mathcal{C}_3^N) & \approx \frac{\alpha}{4}
& R^2_{TL^2_\lambda}(\mathcal{C}_3^N) & \approx \frac{\alpha^2}{N}.
\end{align*}

We have
\begin{align*}
\kappa_{12}^2(L^2;N) & \approx \frac{\alpha N}{4}, & \kappa^2_{13}(L^2;N) & \approx \frac{3\gamma N}{8}, \\
\kappa^2_{12}(\mathrm{OT};N) & \approx \frac{\beta^2 N}{4}, & \kappa^2_{13}(\mathrm{OT};N) & \approx \frac{\beta^2 N^2}{4}, \\
\kappa^2_{12}(TL_\lambda^2;N) & \approx \frac{N}{8\alpha}, & \kappa^2_{13}(TL_\lambda^2;N) & \approx \frac{3N}{4\alpha}, \\
\kappa^2_{23}(L^2;N) & \approx \frac{\gamma N}{8}, \\
\kappa^2_{23}(\mathrm{OT};N) & \approx \frac{\gamma^2 N^2}{8}, \\
\kappa^2_{23}(TL_\lambda^2;N) & \approx \frac{N}{16 \alpha}.
\end{align*}

Finally we can compute $N^*$,
\begin{align*}
N^*_{12}(L^2) & \approx \frac{4}{\alpha}, & N^*_{13}(L^2) & \approx \frac{8}{3\gamma}, & N^*_{23}(L^2) & \approx \frac{8}{\gamma} \\
N^*_{12}(\mathrm{OT}) & \approx \frac{2}{\beta}, & N^*_{13}(\mathrm{OT}) & \approx \frac{2}{\beta}, & N^*_{23}(\mathrm{OT}) & \approx \frac{\sqrt{8}}{\gamma} \\
N^*_{12}(TL^2) & \approx \frac{\alpha}{8}, & N^*_{13}(TL^2) & \approx \frac{4\alpha}{3}, & N^*_{23}(TL^2) & \approx 16 \alpha
\end{align*}
which for $\beta>\frac{\alpha}{2}$, $\beta>\frac{3\gamma}{4}$ and $\gamma<\frac{\sqrt{2}\alpha}{8}$ implies the ordering given Section~\ref{subsec:App:Syn1}.

\section{Numerical Methods \label{sec:app:Numerics}}

In principle any numerical method for OT capable of dealing with an arbitrary cost function can be adapted to compute $TL^p_\lambda$.
Here we describe two numerical methods we used in Section~\ref{sec:App}.

\subsection{Iterative Linear Programming \label{subsec:Num:OR}}

Here we describe the iterative linear programming method of Oberman and Ruan~\cite{oberman15} which we abbreviate OR.
Although this method is not guaranteed to find the minimum in~\eqref{eq:Frame:TLpd} we find it works well in practice and is easier to implement than, for example, methods due to Schmitzer~\cite{schmitzer15} that provably minimize~\eqref{eq:Frame:TLpd} but require a more advanced refinement procedure.
See also~\cite{merigot11} and references therein for a multiscale descent approach.

The linear programming problem restricted to a subset $\mathcal{M}\subseteq\Omega_h^2$ is
\begin{flalign}
\tag{$\mathrm{LP}_h$} \label{eq:Frame:Num:OR:LPh}
\begin{split}
\text{minimize: } & \sum_{(i,j)\in \mathcal{M}} c_\lambda(x_i,x_j;f_h,g_h) \pi_{ij} \text{ over } \pi \\
\text{subject to } & \sum_{i \, : \, (i,j) \in \mathcal{M}} \pi_{ij} = q_j,\sum_{j \, : \, (i,j) \in \mathcal{M}} \pi_{ij} = p_i
\end{split}
\end{flalign}
where $c_\lambda$ is given by~\eqref{eq:Frame:TLpc}.
When $\mathcal{M}=\Omega_h^2$ then the $TL_\lambda^p$ distance between $(f_h,\mu_h)$ and $(g_h,\nu_h)$ is the minimum to the above linear programme.
Furthermore if $\pi_h$ is the minimizer in the $TL_\lambda^p$ distance then it is also the solution to the linear programme in~\eqref{eq:Frame:Num:OR:LPh} for any $\mathcal{M}$ containing the support of $\pi_h$.
That is if one already knows (or can reasonably estimate) the set of nodes $\mathcal{M}$ for which the optimal plan is non-zero then one need only consider the linear programme on $\mathcal{M}$.
This is advantageous when $\mathcal{M}$ is a much smaller set.
Motivated by Proposition~\ref{prop:Frame:Def:ExistTransMap} we expect to be able to write the optimal plan as a map.
This implies whilst $\pi_h$ has $n^2$ unknowns we only expect $n$ of them to be non-zero.

The method proposed by OR is given in Algorithm~\ref{alg:Frame:Num:OR:ORAlg}.
An initial discretisation scale $h_0$ is given and an estimate $\pi_{h_0}$ found for the linear programme~\eqref{eq:Frame:Num:OR:LPh} with $\mathcal{M} = \Omega_{h_0}^2$.
One then iteratively finds $\mathcal{M}_r\subseteq \Omega_{h_r}^2$, where $h_r=\frac{h_{r-1}}{2}$, to be the set of nodes defined by the following refinement procedure.
Find the set of nodes for which $\pi_{h_{r-1}}$ is non-zero, add the neighbouring nodes and then project onto the refined grid $\Omega_{h_r}^2$.
The optimal plan $\pi_{h_r}$ on $\Omega_{h_r}^2$ is then estimated by solving the linear programme~\eqref{eq:Frame:Num:OR:LPh} with $\mathcal{M}=\mathcal{M}_r$.

The grid $\Omega_{h_r}$ will scale as $(2^{rd}h_0^{-1})^2$.
If the linear programme is run $N$ times then at the $r^\text{th}$ step the linear programme has on the order of $2^{rd}h_0^{-1}$ variables.
In particular on the last (and most expensive) step the number of variables is $O(2^{Nd}h_0^{-1})$.
This compares to size $(2^{Nd}h_0^{-1})^2$ if the linear programme was run on the final grid without this refinement procedure.

\begin{algorithm}
\caption{An Iterative Linear Programming Approach~\cite{oberman15} \label{alg:Frame:Num:OR:ORAlg}}
\begin{algorithmic}[1]
\Require functions $f,g\in L^p(\Omega)$, measures $\mu,\nu\in \mathcal{P}(\Omega)$ and parameters $h_0,N$.
\State Set $r=0$.
\Repeat
	\State Define $\mathcal{S}_r=\Omega_{h_r}^2$ where $\Omega_{h_r}$ is the square grid lattice with distances between neighbouring points $h_r$ and discretise functions $f,g$ and measures $\mu,\nu$ on $\Omega_h$. 
	\If{$r=0$}
   		\State Solve~\eqref{eq:Frame:Num:OR:LPh} on $\mathcal{S}_0$ and call the output $\pi_{h_0}$.
   		\Else
		\State Find the set of nodes on $\mathcal{S}_{r-1}$ for which $\pi_{h_{r-1}}$ is non-zero and call the set $\mathcal{K}_{r-1}$.
		\State To $\mathcal{K}_{r-1}$ add all neighbouring nodes and call this set $\mathcal{N}_{r-1}$.
		\State Define $\mathcal{M}_r$ to be the set of nodes on $\mathcal{S}_r$ that are children of nodes in $\mathcal{N}_{r-1}$.
		\State Solve~\eqref{eq:Frame:Num:OR:LPh} restricted to $\mathcal{M}_r$ and call the optimal plan $\pi_{h_r}$.
  	\EndIf
	\State Set $h_{r+1}=\frac{h_r}{2}$ and $r\mapsto r+1$.
\Until{$r=N$}
\Ensure The optimal plan $\pi_{h_{N-1}}$ for~\eqref{eq:Frame:Num:OR:LPh}.
\end{algorithmic}
\end{algorithm}

\subsection{Entropic Regularisation \label{subsec:Num:BCCNP}}

Cuturi, in the context of computing optimal transport, proposed regularizing the minimization in~\eqref{eq:Frame:TLpd} with entropy~\cite{cuturi13}.
This was further developed by Benamou, Carlier, Cuturi, Nenna and Peyr\'{e}~\cite{benamou15}, abbreviate to BCCNP, which is the method we describe here.
Instead of considering the distance $TL^p_\lambda$ we consider
\[ S_\epsilon = \inf_{\pi\in\Pi(\mu,\nu)} \left\{ \sum_{i=1}^n \sum_{j=1}^n c_\lambda(x_i,x_j;f,g) \pi_{ij} - \epsilon H(\pi) \right\} \]
where $H(\pi)= - \sum_{i=1}^n \sum_{j=1}^n \pi_{ij} \log \pi_{ij}$ is the entropy.
In the OT case the distance $S_\epsilon$ is also known as the Sinkhorn distance.
It is a short calculation to show
\[ S_\epsilon = \epsilon \inf_{\pi\in\Pi(\mu,\nu)} \left\{ \mathrm{KL}(\pi|\mathcal{K}) \right\} \]
where $\mathcal{K}_{ij} = \exp\left(-\frac{c_\lambda(x_i,x_j:f,g)}{\epsilon}\right)$ (the exponential is taken pointwise) and $\mathrm{KL}$ is the Kullback-Leibler divergence defined by
\[ \mathrm{KL}(\pi|\mathcal{K}) = \sum_{i=1}^n \sum_{j=1}^n \pi_{ij} \log \left( \frac{\pi_{ij}}{\mathcal{K}_{ij}} \right). \]

It can be shown that the optimal choice of $\pi$ for $S_\epsilon$ can be written in the form $\pi^* = \mathrm{diag}(u) \mathcal{K} \mathrm{diag}(v)$ where $u,v\in\mathbb{R}^n$ are limits, as $r \to \infty$, of the sequence
\[ v^{(0)} = \mathbb{I}, \quad u^{(r)} = \frac{\underline{p}}{\mathcal{K} v^{(r)}}, \quad v^{(r+1)} = \frac{\underline{q}}{\mathcal{K}^\top u^{(r)}} \]
and $\underline{p} = (p_1,\dots,p_n)$, $\underline{q}=(q_1,\dots,q_n)$ (multiplication is the usual matrix-vector multiplication, division is pointwise and $\top$ denotes the matrix transpose).
The algorithm given in~\ref{alg:Frame:Num:BCNNP:BCCNPAlg} is a special case of iterative Bregman projections.

The stopping condition proposed in~\cite{cuturi13} is to let $\pi^{(r)} = \mathrm{diag}(u^{(r)}) \mathcal{K} \mathrm{diag}(v^{(r)})$ then stop when
\[ \left| \frac{\sum_{i,j=1}^n \mathcal{K}_{ij} \pi^{(r)}_{ij} - \epsilon H(\pi^{(r)})}{\sum_{i,j=1}^n \mathcal{K}_{ij} \pi^{(r-1)}_{ij} - \epsilon H(\pi^{(r-1)})} - 1 \right| < 10^{-4}. \]
Note that although as $\epsilon\to 0$ we will recover the unregularised $TL_\lambda^p$ distance we also suffer numerical instability as $\mathcal{K}\to 0$ exponentially in $\epsilon$.

\begin{algorithm}
\caption{An Entropy Regularised Approach~\cite{benamou15,cuturi13} \label{alg:Frame:Num:BCNNP:BCCNPAlg}}
\begin{algorithmic}[1]
\Require discrete functions $f = (f_1,\dots,f_n),g=(g_1,\dots,g_n)$, discrete measures $\mu = \sum_{i=1}^n p_i \delta_{x_i}, \nu = \sum_{j=1}^n q_j \delta_{x_j}$, the parameter $\epsilon$ and a stopping condition.
\State Set $r=0$, $\mathcal{K}=\left(\exp\left(-\frac{c(x_i,x_j;f,g)}{\epsilon}\right)\right)_{ij}$ and $v^{(0)}=\mathbb{I}\in\mathbb{R}^n$.
\Repeat
	\State Let $r\mapsto r+1$,
	\[ v^{(r)} = \frac{\underline{q}}{\mathcal{K}^\top u^{(r-1)}} \quad \text{and} \quad u^{(r)} = \frac{\underline{p}}{\mathcal{K} v^{(r)}} \]
	where $\underline{p} = (p_1,\dots,p_n)$, $\underline{q} = (q_1,\dots, q_n)$.
\Until{Stopping condition has been reached}
\State Set $\pi = \mathrm{diag}(u^{(r)}) \mathcal{K} \mathrm{diag}(v^{(r)})$.
\Ensure An estimate $\pi$ on the optimal plan for $S_\epsilon$ where the accuracy is determined by the stopping condition.
\end{algorithmic}
\end{algorithm}

\small
\bibliographystyle{plain}
\bibliography{references}

\end{document}